\pdfoutput=1

\documentclass[11pt]{article}

\usepackage[]{acl}

\usepackage{times}
\usepackage{latexsym}

\usepackage[T1]{fontenc}

\usepackage[utf8]{inputenc}

\usepackage{microtype}

\usepackage{inconsolata}
\usepackage{graphicx}
\usepackage{multirow}
\usepackage{tabularx}
\usepackage{booktabs}
\usepackage{xspace}
\usepackage{amsmath,amsfonts,amssymb}
\usepackage{float}
\usepackage{slashed}
\usepackage{amsbsy}
\usepackage{amstext}
\usepackage{microtype}
\usepackage{pgfplots}
\pgfplotsset{width=1.0\columnwidth}
\usepackage[multiple]{footmisc}
\usepackage{enumitem}
\usepackage{tabularx}
\usepackage{bbding}

\usepackage{times}
\usepackage{latexsym}
\usepackage{booktabs}
\usepackage{hyperref}
\usepackage[english]{babel}
\usepackage{caption}

\title{Distribution Shifts Are Bottlenecks: Extensive Evaluation for \\ Grounding Language Models to Knowledge Bases}

\author{Yiheng Shu\thanks{* Work performed when the author was graduate student at Nanjing University.} \\
  The Ohio State University \\
  \texttt{shu.251@osu.edu} \\
  \And
  Zhiwei Yu \\
  Microsoft \\
  \texttt{zhiwyu@microsoft.com} \\}

\begin{document}
\maketitle

\begin{abstract}

Grounding language models (LMs) to knowledge bases (KBs) helps to obtain rich and accurate facts. 
However, it remains challenging because of the enormous size, complex structure, and partial observability of KBs. 
One reason is that current benchmarks fail to reflect robustness challenges and fairly evaluate models.
This paper analyzes whether these robustness challenges arise from distribution shifts, including environmental, linguistic, and modal aspects.
This affects the ability of LMs to cope with unseen schema, adapt to language variations, and perform few-shot learning. 
Thus, the paper proposes extensive evaluation protocols and conducts experiments to demonstrate that, despite utilizing our proposed data augmentation method, both advanced small and large language models exhibit poor robustness in these aspects. 
We conclude that current LMs are too fragile to navigate in complex environments due to distribution shifts. This underscores the need for future research focusing on data collection, evaluation protocols, and learning paradigms.\footnote{Code and data are available at \url{https://github.com/yhshu/Distribution-Shifts-for-KBQA}.}

\end{abstract}

\section{Introduction}
\label{sec:introduction}

Language models (LMs), such as BERT \cite{devlin19bert}, T5 \cite{raffel20exploring}, and the GPT series \cite{ouyang2022training,gpt4}, have demonstrated impressive capabilities in understanding and generating languages, highlighting the potential for artificial general intelligence (AGI). 
However, a major obstacle to achieving this goal is that LMs mainly built on natural languages are not yet well-grounded to real-world environments, such as knowledge base (KB), an environment of enormous size, complex structure, and only partially observable to LM.


\begin{figure*}
    \centering
    \includegraphics[width=\textwidth]{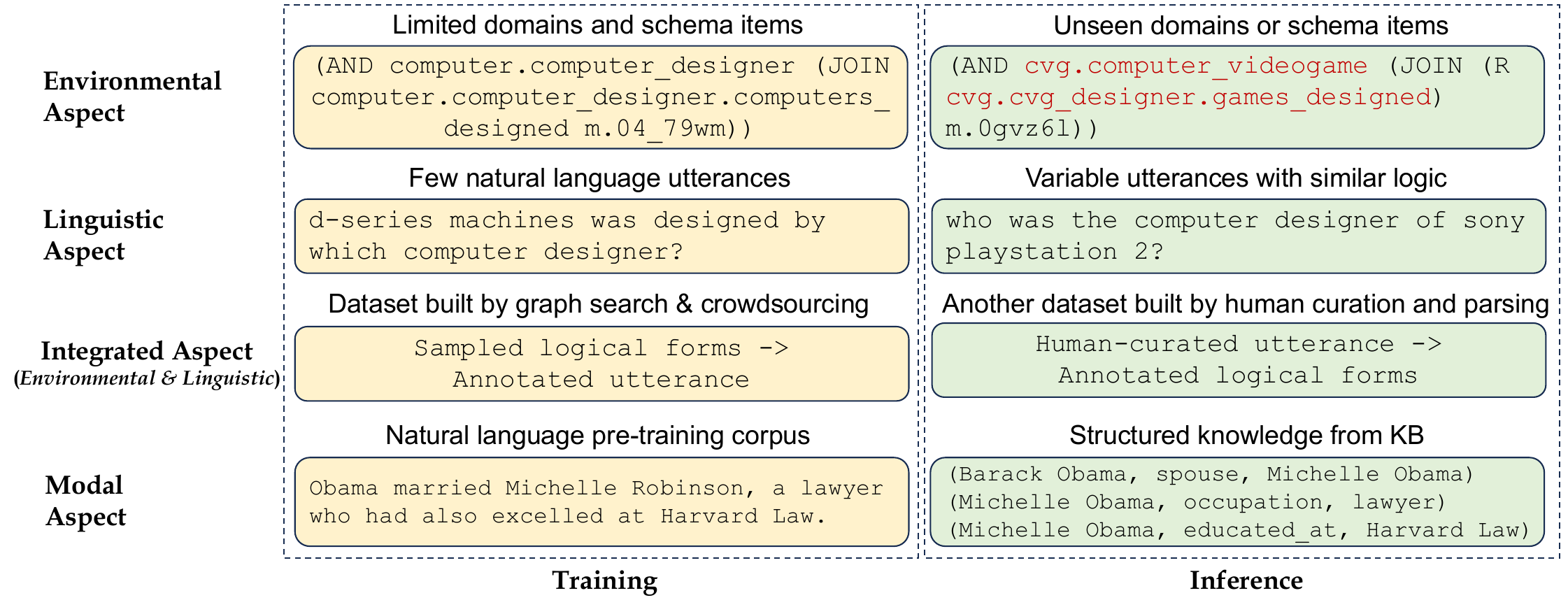}
    \caption{Distribution shifts during grounding in the case of KBQA. Training and inference using LM face completely different data distributions. We call for these shifts to be reflected in the evaluation protocols.}
    \label{fig:kbqa robustness}
\end{figure*}

Though LMs are highly skilled at natural language question answering (QA) today, the task of Knowledge Base Question Answering (KBQA) aims to parse natural language queries into formal queries on KBs, such as Freebase \cite{bollacker08freebase} and Wikidata \cite{vrandecic14wikidata}.
The significance of this task lies in building language agents on complex environments \cite{su2023language}, rather than merely recalling answers from the LM's stored knowledge.

Now, numerous LM-driven models \cite{das21case,hu2022logical} continue to achieve higher F1/Hits@1 scores on KBQA benchmarks.
However, achieving higher scores does not necessarily guarantee the development of robust and dependable models.
We still need to ask whether such improvements apply to extensive scenarios, as benchmarks almost always create questions via crowdsourcing and evaluate with simplistic metrics (Table \ref{tab:KBQA benchmark}).
These benchmarks may not fully represent the diverse scenarios encountered in real-world applications, which raises concerns about the robustness of LM-driven models. 
Thus, our research aims to bridge this gap by exploring the limitations of current KBQA benchmarks and proposing more comprehensive evaluation protocols.

To achieve this goal, we need to grasp the key factors in robustness.
For modern deep learning systems, the amount of training data could be extremely rich, but \textit{robustness is closely related to data distribution} \cite{hendrycks20themany}.
In the general area of natural language processing (NLP), large-scale corpora have been collected and used for effective training \cite{touvron23Llama2}.
However, real-world environments are rarely so accommodating, e.g., large KBs contain complex structures and schema items, and building a large-scale and representative corpus is quite challenging.
The problem of inconsistent data distribution during training and inference, i.e., \textbf{distribution shifts} as shown in Figure \ref{fig:kbqa robustness}, may negatively impact the performance and robustness of grounded LMs.




This paper analyzes the distribution shifts from multiple aspects to understand both KBQA benchmarking and LM grounding better.
We review existing works and identify several challenges.
1) \textbf{Environmental aspect}: generalization to unseen domains at the schema\footnote{Schema denotes rdfs:Class (class) and rdf:Property (relation) here.} level \cite{gu21beyond};
2) \textbf{Linguistic aspect}: adaptation to paraphrases featuring diverse language variations \cite{su16graphq};
3) \textbf{Integrated aspect}: transferability across datasets with both novel schema items and linguistic styles \cite{cao22program}; 
4) \textbf{Modal aspect}: few-shot learning ability of large language models (LLMs) for the unfamiliar KB modality \cite{li23fewshot}.
These challenges guide us to build more rigorous evaluation protocols for multiple aspects (\S\ref{subsec:evaluation protocols}).

To measure the impact of distribution shifts, we conduct extensive experiments under the proposed evaluation protocols.
To present a fair evaluation under such rigorous protocols, we implement a data augmentation method for evaluated LMs and a retrieval augmentation method for evaluated LLMs (\S\ref{sec:approach}).
Our findings reveal that even when employed with such methods and the highest EM scores are achieved on the GrailQA benchmark \cite{gu21beyond}, \textit{advanced small and large LMs still fall short of effectively tackling the majority of these challenges}.
A striking example is the large difference between the best practice without WebQSP \cite{yih16thevalue} fine-tuning (F1 43.0\%) compared to the fine-tuned state-of-the-art (F1 79.6\%) (\S\ref{sec:analysis}), suggesting the weak robustness of LM-driven KBQA models on an unseen dataset.
Such negative observations highlight an urgent need for future research in data collection methodologies and LM learning paradigms.
Meanwhile, we expect our evaluation protocols to provide a reference for future benchmark construction, developing metrics that consider robustness.

Our contributions include 1) A systematic \textbf{analysis} of the robustness challenges of grounded LMs and a strong advocacy of a multi-aspect \textbf{evaluation protocol} for KBQA. 
2) Extensive \textbf{experiments} quantitatively unveil the existence of these challenges and the vulnerability of both small and large LMs.
3) \textbf{Insights} into improving the robustness of grounded LMs, including data collection and learning paradigms.

\section{Related Work}
\label{sec:related work}


Compared to existing research, the necessity of this study is threefold. 
First, while LMs are commonly evaluated on natural language tasks \cite{hupkes22state,patel22revisiting}, the complexity increases significantly when these models are applied to environments like KBs, where data is structured rather than purely unstructured natural language \cite{liu23agentbench}. 
Second, the evaluation protocols used in KBQA benchmarks tend to be uniform, leading to an insufficient evaluation of model robustness \cite{gu21beyond}. 
Finally, recent reviews in this field \cite{lan21complex,gu22knowledge} have largely overlooked the advancements in the development and application of LMs, particularly LLMs.



\begin{table*}[htb]
\centering
\small
\resizebox{\textwidth}{!}{
\begin{tabular}{lcrllccc}
\toprule
\textbf{Benchmark} & \textbf{KB} & \textbf{Size} & \textbf{LF} &  \textbf{Generalization} & \textbf{Paraphrases} & \textbf{Metrics} \\
\toprule
WebQuestions~\cite{berant13semantic} & Freebase & 5,810 & N/A & i.i.d. & \XSolidBrush & F1 \\
SimpleQuestions~\cite{bordes15large} & Freebase & 108,442 & N/A & i.i.d. & \XSolidBrush & Acc \\
WebQuestionsSP \cite{yih16thevalue} & Freebase & 4,737 & SPARQL & i.i.d. & \XSolidBrush & F1  \\
GraphQuestions~\cite{su16graphq}  & Freebase & 5,166 & Graph query  & comp.+zero & \Checkmark & F1  \\
LC-QuAD~\cite{trivedi17lcquad} & DBpedia & 5,000 & SPARQL &  i.i.d. & \XSolidBrush & F1 \\
CWQ \cite{talmor2018web} & Freebase & 34,689 & SPARQL &  i.i.d. & \XSolidBrush & Hits \\
LC-QuAD 2.0~\cite{du19lcquad2} & Wikidata & 30,000 & SPARQL &  i.i.d. & \Checkmark & F1 \\
SQB~\cite{wu19learning} & Freebase & 108,443 & N/A & i.i.d.+zero & \XSolidBrush & Acc \\
CFQ~\cite{keysers20measuring} & Freebase & 239,357 & SPARQL &  comp & \XSolidBrush & Acc \\
GrailQA \cite{gu21beyond} & Freebase & 64,331 & S-expression &  i.i.d.+comp.+zero & \Checkmark & EM, F1 \\
KQA Pro~\cite{cao22kqa} & Wikidata & 117,970 & KoPL &  i.i.d. & \XSolidBrush & Acc \\
QALD series~\cite{Perevalov22qald9} & DBpedia & 558 & SPARQL &  comp. & \XSolidBrush & F1 \\
\bottomrule
\end{tabular}
}   
\caption{
Selected KBQA benchmarks. 
LF denotes logical forms. 
    Generalization settings follow \citet{gu21beyond}. 
    \textit{i.i.d.} denotes that the schema distribution in the test set is the same as the training set. 
    \textit{comp.} and \textit{zero} denote compositional and zero-shot generalization, respectively. 
    Paraphrases are questions containing the same semantics (machine-generated paraphrases are not included).
    \textit{Acc} denotes accuracy. \textit{EM} denotes exact match.
    }
    \label{tab:KBQA benchmark}
\end{table*}

\section{Challenges from Distribution Shifts}
\label{sec:robustness challenge}

In this paper, the \textbf{robustness} of a KBQA model refers to its ability to adapt to various natural language inputs and maintain consistent performance when data distribution shifts.
Due to the distribution shifts between the training corpus of LMs and KB environments, grounding LMs face robustness challenges from environmental, linguistic, and modal aspects.

\subsection{Environmental Aspect}
\label{subsec:schema-level generalization}

A primary environmental challenge is \textbf{schema-level generalization}. 
The RDF Schema provides a data-modeling vocabulary crucial for querying a KB. 
Table~\ref{tab:KBQA benchmark} indicates most KBQA benchmarks assume a consistent schema distribution between training and testing. 
However, this often does not hold in large KBs with numerous schema items. 
Few benchmarks, like the reconfigured SimpleQuestions-Balance dataset \cite{wu19learning} and GrailQA \cite{gu21beyond}, address non-i.i.d. schema items and varying levels of schema-level generalization. 
GraphQuestions \cite{su16graphq} provides a stringent test with seldom-seen schema items in training. 
Given real-world non-i.i.d. complexities, these datasets better represent the practical generalization challenges. 
Yet, despite progress \cite{shu22tiara,gu2022don}, compositional and zero-shot generalization are far from solved.
We stress that this challenge applies to the overall KBQA semantic parsing process, especially to important subtasks such as relation linking.

\subsection{Linguistic Aspect}
\label{subsec:paraphrase adaptation}

Various natural language expressions make question understanding challenging for KBQA models.
One common way this variety shows up is through paraphrasing. 
In this paper, a paraphrase set denotes different ways to express the same logical form, as illustrated in Table~\ref{tab:paraphrase example}.
\textbf{Paraphrase adaptation} is an intuitive form of the ability to comprehend variable language expressions. 
It could be measured by whether a model is able to accurately answer paraphrased questions that this model has already answered correctly before. 
Unfortunately, as shown in Table \ref{tab:KBQA benchmark}, many KBQA benchmarks do not account for paraphrasing with only one utterance for each logical form.
Exceptionally, some datasets \cite{su16graphq,du19lcquad2,gu21beyond} are based on automatically generated logical forms and include multiple natural language expressions for the same logical form (template).
These data characteristics highlight the difficulties in adapting to paraphrased questions.

\subsection{Integrated Aspect}

Evaluating KBQA benchmarks often hinges on a single dataset, thereby complicating the task of ascertaining the model performance consistency across novel scenarios. 
This form of robustness, termed as \textbf{cross-dataset transfer} in this paper, combines both the environmental and linguistic aspects discussed earlier and is more difficult to achieve. 
This is because construction methods vary across datasets, as do schema distributions and natural language expressions.
Specifically, KBQA dataset construction generally falls into two distinct categories:
1) Graph Search and Crowdsourcing: in this approach, logical forms or triples are initially extracted from a KB, where structures or operators of logical form are usually finite. Subsequently, they are converted into natural language utterances through crowdsourcing methods \cite{bordes15large,trivedi17lcquad}.
2) Human Curation and Parsing: logical forms are labeled directly from human-provided utterances \cite{berant13semantic,Perevalov22qald9}.
Existing works \cite{gu21beyond,cao22program} suggest that models pre-trained on large-scale datasets can adapt reasonably well to other target datasets, such as WebQSP \cite{yih16thevalue}. 
However, the necessity for fine-tuning these pre-trained models on the intended target dataset remains imperative for achieving optimal performance. 
Despite the advantages offered by pre-training on expansive KBQA datasets, models still encounter challenges in transferring directly to previously unseen target datasets while sustaining high performance.

\subsection{Modal Aspect}
\label{subsec:modal aspect}

Aside from considering environmental and linguistic factors, focusing on the modal aspect is also crucial.
Recently, LLMs like GPT series \cite{gpt4} have demonstrated exceptional capabilities across a variety of tasks, outperforming smaller yet potent LMs such as BERT \cite{devlin19bert} and T5 \cite{raffel20exploring}. 
Despite these advancements, these LLMs face substantial challenges when interacting with environments. 
One notable issue is they predominantly rely on an in-context learning paradigm as opposed to fine-tuning, as a trade-off between computational cost and model efficiency.
In comparison to fine-tuning, in-context learning offers the advantage of reduced training costs but at the expense of being forced to reason over the \textbf{unfamiliar modality}.
Distribution shifts between natural language pre-training and reasoning over structured knowledge contexts could lead to poor performance.
For instance, a discernible performance gap exists between KBQA models that employ in-context learning with Codex \cite{chen21evaluating} and those built on fine-tuned LMs \cite{gu2022don,li23fewshot}.
However, the empirical specifics of this difference are not yet clear, leaving us with an inadequate understanding of the limitations of in-context learning and ways to improve grounding with LLMs.

\section{Evaluation Protocols}
\label{subsec:evaluation protocols}

Regarding these challenges, we introduce extensive protocols for evaluating LMs in several aspects overlooked by current benchmarks.

\subsection{Evaluating Environmental Aspect}

To set the environmental schema-level generalization scenario, we use GrailQA \cite{gu21beyond} and GraphQuestions \cite{su16graphq} datasets.
GrailQA contains three generalization levels: i.i.d. (25\%), compositional (25\%), and zero-shot (50\%). 
GraphQuestions has no seen relations in the test set.
We also use SimpleQuestions-Balance (SQB) \cite{wu19learning} for the relation linking task (an important KBQA sub-task), where 50\% of the samples contain unseen relations.

\subsection{Evaluating Linguistic Aspect}
\label{subsec:evaluation protocol linguistic}

To set a paraphrase adaptation scenario, we use GrailQA \cite{gu21beyond} and GraphQuestions \cite{su16graphq} datasets.
To evaluate adaptability to paraphrases (\S\ref{subsec:paraphrase adaptation}), we propose a new metric, the standard deviation (std) of EM/F1 for questions of each logical form template. 
As shown in Equation~\ref{equ:f1 std}, suppose there are $n$ sets of paraphrases in the dataset, each set of paraphrases corresponds to a logical form template with $m$ natural language expressions, and the F1 score obtained by the KBQA model on the $j$-th question of the $i$-th set of paraphrases is $F1_{i,j}$. 
The metric $Std_{F1}$ first calculates the standard deviation of the F1 scores obtained by the model on the $m$ questions for each set of paraphrases and then calculates the average of the $n$ standard deviations.
This metric is used to measure the robustness of the model to different representations of the same semantics, i.e., whether it can cope with diverse natural language expressions. 
A lower standard deviation indicates that the model is more adaptive to different expressions. 
$Std_{EM}$ is calculated in the same way.

\begin{equation}
Std_{F1} =  \frac{1}{n} \sum_{i=1}^{n} \sqrt{ \left(\frac{\sum_{j=1}^{m}(F1_{i,j} - \bar{F1_i})^2}{m}\right)}
\label{equ:f1 std}
\end{equation}

\subsection{Evaluating Integrated Aspect}
To emulate a real-world scenario with unknown schema and linguistic distribution for the integrated aspect, we evaluate the performance of pre-trained models on the unseen human-curated WebQSP \cite{yih16thevalue} dataset, where the questions are derived from search logs and are more realistic. 
This is a scenario where the distribution changes significantly, as most benchmarks create questions by sampling logical forms and annotating natural language questions via crowdsourcing \cite{lan21complex}, where questions are confined to sampled logical forms.

\subsection{Evaluating Modal Aspect}
To test the capability of the in-context learning paradigm to inference from the KB modality rather than from pure texts, we retrieve structured KB contexts as prompt to evaluate the LLM without particular fine-tuning on KB (\S\ref{subsec:retrieval augmentation for LLMs}). 
KBs are structured and expansive, but in this case, LLM can only encode a portion of linearized KB contexts.

\section{Experiments}
\label{sec: experiments}

\subsection{Augmentation Approach}
\label{sec:approach}

To ensure fair evaluation and fully harness the capabilities of LMs under our extensive and rigorous evaluation protocols, we suggest two strategies to counteract distribution shifts: data augmentation and retrieval augmentation.


\subsubsection{Data Augmentation for LMs}

Off-the-shelf datasets of limited size may make LM easily overfitted and not adaptable to large KBs.
To address the problem that many domains in the KB are often not collected as training data,
we propose a data augmentation method named \textbf{G}raph se\textbf{A}rch and quest\textbf{I}on generatio\textbf{N} (\textbf{GAIN}). 
Some data augmentation or question generation models \cite{bi20knowledge,guo22dsm} are only evaluated by the qualify of generated sentences rather than evaluated by the QA task, but GAIN directly serves our KBQA evaluations.
Besides, compared to the previous work \cite{hu2019question} that only considers generating questions for triples to help KBQA, GAIN applies to KBQA corresponding to both logical forms and triples.
GAIN scales data volume and distribution through four steps:
1) Graph search: Sampling logical forms or triples from arbitrary domains in the KB without being restricted to any particular KBQA dataset.
2) Training question generator: learning to convert logical forms or triples into natural language questions on existing KBQA datasets.
3) Verbalization: Using the question generator from step 2 to verbalize sampled logical forms or triples from step 1, thus creating synthetic questions.
4) Training data expansion: Before fine-tuning any neural models on KBQA datasets, GAIN-synthetic data can be used to train these models or to expand the corpus of in-context samples for LLMs.
That is, as a data augmentation method, GAIN is not a KBQA model, but it is used to augment a base KBQA model.

\subsubsection{Retrieval Augmentation for LLMs}
\label{subsec:retrieval augmentation for LLMs}

As the trade-off between cost and effectiveness, we experiment with the prevalent in-context learning paradigm but attempt to improve the quality of in-context samples. 
We use advanced retrieval methods based on smaller LMs as plug-ins to augment the LLM, similar to the SuperICL approach \cite{xu2023small}. 
Specifically, our steps to generate an LLM prompt for each question include the following.
1) Given an input question, we retrieve $k$ questions ($k$-shot) with BM25 \cite{robertson09probabilistic} from the corpus (the combination of KBQA training set and the GAIN-synthetic dataset). 
2) The role of retrieval augmentation for KB environments has been shown by fine-tuned LMs \cite{shu22tiara}.
To assist with grounding LLM, we retrieve KB contexts with off-the-shelf retrievers for $k$ samples and the input question.\footnote{The prompt example is demonstrated in Appendix \ref{sec:examples of llm prompt}.}



\subsection{Setup}
\label{subsec: setup}

\paragraph{Data}

All experiments use S-expression \cite{gu21beyond} as the logical form due to its clear and concise structure. 
Entity linking results are taken from TIARA \cite{shu22tiara} for GrailQA and WebQSP, and ArcaneQA \cite{gu22arcaneqa} for GraphQuestions, because of their public availability and performance.

\paragraph{Model}
Compared models are mainly selected from the leaderboard.\footnote{\url{https://dki-lab.github.io/GrailQA/}}
The performances are taken from their papers.
For the relation linking task on SQB, we use BERT \cite{devlin19bert} as the base model for GAIN.
For KBQA tasks, we use the open-source advanced model TIARA \cite{shu22tiara} as the base model for GAIN, due to its strong performance on zero-shot schema items.\footnote{Pangu \cite{gu2022don} also uses entity linking results from TIARA.}
TIARA is composed of multi-grained retrievers and a generator, with the retrievers providing KB contexts\footnote{Entities, exemplary logical forms, and schema items are retrieved.} for the generator.
The term ``TIARA+GAIN'' represents a model (both the retrievers and the generator) that is first tuned using GAIN synthetic data and subsequently fine-tuned on a target dataset.
For LLM evaluation in the modal aspect, we use the \texttt{gpt-3.5-turbo-0613}\footnote{\url{https://platform.openai.com/docs/models}} model, and the few-shot contexts are retrieved from the combination of GrailQA training set and synthetic dataset using the TIARA+GAIN retrievers.

\paragraph{Metrics}
Following previous works, we use Exact Match (EM), F1, and Hits@1 to measure the performance of KBQA models.
We also use the std of EM/F1 to measure the adaptability to paraphrases (\S\ref{subsec:evaluation protocol linguistic}).

\begin{table*}[htb]
\small
\centering
\begin{tabular}{lcccccccc}
\toprule
&\multicolumn{2}{c}{\textbf{Overall}}   &\multicolumn{2}{c}{\textbf{I.I.D.}}  &\multicolumn{2}{c}{\textbf{Compositional}}    &\multicolumn{2}{c}{\textbf{Zero-shot}} \\
 \cmidrule(lr){2-3} \cmidrule(lr){4-5} \cmidrule(lr){6-7} \cmidrule(lr){8-9}
\textbf{Model on GrailQA Test Set} & \textbf{EM} & \textbf{F1} & \textbf{EM} & \textbf{F1} & \textbf{EM} & \textbf{F1} & \textbf{EM} & \textbf{F1} \\ 
\midrule
\multicolumn{9}{c}{\textit{Fine-tuned Models}} \\
BERT + Ranking~\citep{gu21beyond}      & 50.6  & 58.0 & 59.9  & 67.0 & 45.5  &  53.9  & 48.6  & 55.7  \\
RnG-KBQA~\citep{ye21rng} & 68.8 & 74.4 & 86.2 & 89.0 & 63.8 & 71.2 & 63.0 & 69.2 \\
TIARA (T5-base)~\citep{shu22tiara} & 73.0 & 78.5 & 87.8 & 90.6 & 69.2 & 76.5 & 68.0 & 73.9 \\ 
DecAF (FiD-3B)~\citep{yu22decaf} & 68.4 & 78.8 & 84.8 & 89.9 & 73.4 & 81.8 & 58.6 & 72.3 \\  
Pangu (BERT-base)~\citep{gu2022don} & 73.7 & 79.9 & 82.6 & 87.1 & 74.9 & 81.2 & 69.1 & 76.1 \\
Pangu (T5-large)~\citep{gu2022don} & 74.8 & 81.4 & 82.5 & 87.3 & \textbf{75.2} & \textbf{82.2} & 71.0 & 78.4 \\
Pangu (T5-3B)~\citep{gu2022don} & 75.4 & \textbf{81.7} & 84.4 & 88.8 & 74.6 & 81.5 & 71.6 & \textbf{78.5} \\ 
\midrule 
\multicolumn{9}{c}{\textit{Codex-driven Models}} \\
KB-BINDER (6)-R ~\citep{li23fewshot} & 53.2 & 58.5 & 72.5 & 77.4 & 51.8  & 58.3 & 45.0 & 49.9 \\
Pangu (Codex)~\citep{gu2022don} & 56.4 & 65.0 & 67.5 & 73.7 & 58.2 & 64.9 & 50.7 & 61.1 \\
\midrule
\multicolumn{9}{c}{\textit{GAIN-augmented Models}} \\
TIARA + \textbf{GAIN} (T5-base) & 75.1 & 80.6 & 88.3 & 91.0 & 73.0 & 79.6 & 69.9 & 76.4 \\
TIARA + \textbf{GAIN} (T5-3B) & \textbf{76.3} & 81.5 & \textbf{88.5} & \textbf{91.2} & 73.7 & 80.0 & \textbf{71.8} & 77.8 \\
GPT-3.5-turbo (5-shot) & 66.6 & 71.4 & 82.7 & 85.3 & 60.5 & 66.3 & 61.9 & 67.2 \\ 
\bottomrule
\end{tabular}
\caption{EM and F1 scores (\%) on the hidden test set of GrailQA. 
}
\label{tab:grailqa test results}
\end{table*}

\subsection{Implementation Details}

We use a machine with an NVIDIA A100 GPU and up to 504GB of RAM. Models are implemented by PyTorch \cite{paszke19pytorch} and Hugging Face.\footnote{\url{https://huggingface.co/}}
TIARA+GAIN (T5-3B) takes about 100 hours to train the logical form generator on the synthetic dataset.

\paragraph{Model Training}
1) For question generation, we fine-tune the T5-base model \cite{raffel20exploring} to convert S-expression or triple to natural language questions.
We set the beam size to 10, the learning rate to 3e-5, the number of epochs to 10, and the batch size to 8.
2) The training of the TIARA model \cite{shu22tiara} follows its original settings, including the setting of hyperparameters and the calculation of metrics.
Note that Hits@1 on TIARA is obtained by randomly selecting one answer for each question 100 times.
Both the schema retriever and generator of TIARA are pre-trained on synthetic data and then fine-tuned on KBQA datasets.
Since GraphQuestions has no official training-valid split, we randomly take 200 questions from the original training set as the valid set.
3) We use BERT-base-uncased \cite{devlin19bert} to rank candidate relations for SQB, and the input form is the same as the schema retriever of TIARA.
We set the learning rate to 3e-5, the batch size to 256, and the max number of epochs to 3 with early stopping.


\paragraph{Data Augmentation}

The statistics of GAIN-synthetic datasets for both logical forms and triples are shown in Table~\ref{tab:statistics for synthetic of LF} and \ref{tab:statistics for synthetic of triple}.\footnote{Details of synthetic data are shown in Appendix~\ref{sec:examples of synthetic data}.}
Note that the sampling of the GAIN method is not limited to the scale of the synthetic data we use here.

\section{Analysis}
\label{sec:analysis}

\begin{table}[htb]
\small
\centering
\resizebox{\linewidth}{!}{
\begin{tabular}{lcc}
\toprule
   \textbf{Model on GraphQuestions}  & \textbf{F1}($\uparrow$) & \textbf{Std}($\downarrow$) \\ \midrule
   \multicolumn{3}{c}{\textit{GraphQuestions on Freebase 2013-07}} \\
    UDepLambda~\cite{reddy17universal} & 17.7 & - \\
    PARA4QA~\cite{dong17learning} & 20.4 & - \\
    SPARQA~\cite{sun20sparqa} & 21.5 & - \\
   BERT + Ranking~\cite{gu21beyond} & 25.0 & - \\
   ArcaneQA~\cite{gu22arcaneqa}  & 31.8  & - \\ 
   TIARA$^\clubsuit$ (T5-base)~\cite{shu22tiara} &  37.9 &  \textbf{0.141} \\
   KB-BINDER (6)~\cite{li23fewshot} & 39.5 & - \\
   TIARA + \textbf{GAIN} (T5-base) & 45.5 &  0.153 \\ 
   TIARA + \textbf{GAIN} (T5-3B) & \textbf{48.7} &  0.180 \\ 
   
   \midrule
   \multicolumn{3}{c}{\textit{GraphQuestions on Freebase 2015-08-09}} \\
    BERT + Ranking~\cite{gu21beyond} & 27.0 & - \\
   ArcaneQA~\cite{gu22arcaneqa}  & 34.3 & - \\ 
   TIARA$^\clubsuit$ (T5-base)~\cite{shu22tiara} &  41.2 & \textbf{0.157} \\
   Pangu (Codex)~\cite{gu2022don} & 44.3 & - \\
   Pangu (T5-3B)~\cite{gu2022don} & \textbf{62.2} & - \\
   TIARA + \textbf{GAIN} (T5-base) & 49.5 &  0.170 \\ 
   TIARA + \textbf{GAIN} (T5-3B) & 53.0 &  0.200 \\ 
   \bottomrule
\end{tabular}
}
\caption{F1 scores (\%) and average standard deviation (std) of F1 scores for each set of paraphrases on the test set of GraphQuestions. The setting for Freebase 2015-08-09 is described by \citet{gu22arcaneqa}. $^\clubsuit$ denotes our replication results.}
\label{tab:graphq test results}
\end{table}

\begin{table*}[htb]
\small
\centering
\begin{tabular}{lccccccccc}
\toprule
&\multicolumn{3}{c}{\textbf{Overall}}   &\multicolumn{3}{c}{\textbf{Seen}}  &\multicolumn{3}{c}{\textbf{Unseen}} \\
 \cmidrule(lr){2-4} \cmidrule(lr){5-7} \cmidrule(lr){8-10}
\textbf{Model on SimpleQuestions-Balance} & \textbf{1} & \textbf{5} & \textbf{10} &  \textbf{1} & \textbf{5} & \textbf{10} & \textbf{1} & \textbf{5} & \textbf{10} \\ \midrule
   HR-BiLSTM~\cite{wu19learning} & 63.3 & - & - & \textbf{93.5} & - & - & 33.0 & - & - \\
   Adversarial-Adapter~\cite{wu19learning}  & 84.9 & - & - & 92.6 & - & - & 77.1 & - & - \\ \midrule
   BERT-base & 83.7 & 95.0 & 96.9 & 85.8 & 95.0 & 96.0 & 81.5 & 95.1 & 97.8 \\ 
   BERT-base + \textbf{GAIN} & \textbf{88.4} & \textbf{96.0} & \textbf{97.3} & 87.8 & 95.4 & 96.3 & \textbf{89.1} & \textbf{96.7} & \textbf{98.4} \\ 
   \bottomrule 
\end{tabular}
\caption{Hits@$k$ (1, 5, 10) scores (\%) for relation linking on the test set of SimpleQuestions-Balance, including seen and unseen relations.}
\label{tab:sqb relation linking}
\end{table*}

\begin{table*}[htb]
\small
\centering
\begin{tabular}{lcccccccc}
\toprule
& \multicolumn{2}{c}{\textbf{Std Overall}} & \multicolumn{2}{c}{\textbf{Std I.I.D.}}  &\multicolumn{2}{c}{\textbf{Std Compositional}} &\multicolumn{2}{c}{\textbf{Std Zero-shot}} \\
 \cmidrule(lr){2-3} \cmidrule(lr){4-5} \cmidrule(lr){6-7} \cmidrule(lr){8-9}
\textbf{Model on GrailQA Valid Set} & \textbf{EM}($\downarrow$) & \textbf{F1}($\downarrow$) & \textbf{EM} & \textbf{F1} & \textbf{EM} & \textbf{F1} & \textbf{EM} & \textbf{F1}  \\ 
\midrule
TIARA (T5-base)~\cite{shu22tiara} & 0.079 & 0.066 & 0.021 & 0.017 & 0.211 & 0.203 & 0.222 & 0.181 \\  
TIARA + \textbf{GAIN} (T5-base) & 0.077 & 0.061 & 0.020 & 0.016 & 0.215 & 0.198 & 0.218 & 0.160 \\ 
TIARA + \textbf{GAIN} (T5-3B) & \textbf{0.075} & \textbf{0.058} & \textbf{0.020} & \textbf{0.016} & \textbf{0.196} & \textbf{0.180} & \textbf{0.212} & \textbf{0.155} \\ 
GPT-3.5-turbo (5-shot) & 0.093&0.091&0.027&0.023&0.272&0.281&0.251&0.247 \\
\bottomrule
\end{tabular}
\caption{Average \textbf{standard deviation} of EM and F1 scores for each set of paraphrases on the GrailQA valid set. }
\label{tab:grailqa validation std results}
\end{table*}

We report and analyze the experimental results in this section for each aspect.

\subsection{Analysis of Environmental Aspect}

\paragraph{Effectiveness of Synthesis and Scaling Up}

As shown in Tables~\ref{tab:grailqa test results} and \ref{tab:graphq test results}, the models perform significantly better on i.i.d. than compositional and zero-shot generalization, with the zero-shot partition being the most challenging. 
TIARA+GAIN (T5-base) improves 2.5 zero-shot F1 points compared to TIARA (T5-base).
Besides, an increased number of model parameters, combined with richer data from GAIN, significantly enhance the generalization capabilities of T5 models.
TIARA+GAIN (T5-3B) further improves 1.4 zero-shot F1 points compared to its T5-base version.
TIARA+GAIN achieves the highest EM scores, including that on zero-shot scenes.
It demonstrates promising ideas for further improving LM generalization capabilities, i.e., the positive effect of synthetic data and parametric scales on training LMs.

\paragraph{Fine-tuning Better Than Few-shot Learning}

However, it is important to note that fine-tuned models consistently outperform few-shot learning models, regardless of whether the schema is seen or not.
Given the training and inference costs of LLMs, their performance has yet to show any superiority in this task.


\subsection{Analysis of Linguistic Aspect}

\paragraph{Improvements Are Linguistic Biased}
We calculate the standard deviation (std) of EM or F1 in the dev/test set, as shown in Equation \ref{equ:f1 std}.
For GrailQA, the std of EM and F1 decreases with the application of GAIN or an increase in model size, i.e., F1/EM and the std of F1/EM are both better, as shown in Table \ref{tab:grailqa test results} and \ref{tab:grailqa validation std results}.
However, in the case of more challenging GraphQuestions, GAIN significantly improves the F1 by 8.3 points but also results in a larger std (0.170 compared to 0.157), as shown in Table~\ref{tab:graphq test results}. 
It suggests that improving paraphrase adaptation using GAIN is more difficult when the base model (TIARA, T5-base, with only 37.9\% F1) still struggles to address most of the dataset.
Consequently, the performance gains observed on the KBQA benchmark may not necessarily reflect a deeper understanding of linguistic complexities, but they could simply render the model more sensitive to specific phrases.
Strategies for deeper decomposition and understanding \cite{hu21edg,huang23question} of the questions may be needed to mitigate this challenge.

\begin{table}[t]
\small
\centering
\begin{tabular}{lcc}
\toprule
   \textbf{Model on WebQSP} & \textbf{F1} & \textbf{Hits@1} \\ \midrule
    TIARA$^\clubsuit$  (T5-base) \cite{shu22tiara} & 28.5 & 27.6 \\
    TIARA*$^\clubsuit$ (T5-base) \cite{shu22tiara} & 33.5 & 31.5 \\
    BERT + Ranking*~\cite{gu21beyond} & \textbf{43.0} & -    \\ \midrule
    TIARA + \textbf{GAIN} (T5-base)  & 29.1 & 28.2 \\ 
    TIARA + \textbf{GAIN} (T5-3B)    & 29.8 & 28.7 \\
    TIARA* + \textbf{GAIN} (T5-base) & 33.9 & 31.8 \\
    TIARA* + \textbf{GAIN} (T5-3B)   & 34.5 & 32.3 \\
    \bottomrule
\end{tabular}
\caption{F1 and Hits@1 scores (\%) on WebQSP without fine-tuning on it. 
All models are trained on large-scale GrailQA.
* denotes using oracle entity annotations. 
$^\clubsuit$ denotes our replication results.
}
\label{tab:webqsp dataset transfer}
\end{table}

\begin{table*}[htb]
\small
\centering
\begin{tabular}{lcccccccc}
\toprule
& \multicolumn{2}{c}{\textbf{Overall}} & \multicolumn{2}{c}{\textbf{I.I.D.}}  &\multicolumn{2}{c}{\textbf{Compositional}} &\multicolumn{2}{c}{\textbf{Zero-shot}} \\
 \cmidrule(lr){2-3} \cmidrule(lr){4-5} \cmidrule(lr){6-7} \cmidrule(lr){8-9}
\textbf{Model on GrailQA Valid Set} & \textbf{EM} & \textbf{F1} & \textbf{EM} & \textbf{F1} & \textbf{EM} & \textbf{F1} & \textbf{EM} & \textbf{F1} \\ 
\midrule
\multicolumn{1}{l}{BERT + Ranking~\citep{gu21beyond}}  & 51.0 & 58.4 & 58.6 & 66.1 & 40.9 & 48.1 & 51.8 & 59.2 \\
TIARA ELF only \cite{shu22tiara} & 67.2 & 72.9 & 72.8 & 76.7 & 55.3 & 60.7 & 69.7 & 76.3  \\
RnG-KBQA~\cite{ye21rng}  & 71.4 & 76.8 & 86.7 & 89.0 & 61.7 & 68.9 & 68.8 & 74.7  \\
DecAF (FiD-3B)~\cite{yu22decaf}  & - & 81.4 & - & 89.7 & - & \textbf{80.1} & - & 78.4 \\ 
TIARA (T5-base)~\cite{shu22tiara} & 75.3 & 81.9 & 88.4 & 91.2 & 66.4 & 74.8 & 73.3 & 80.7 \\ 
Pangu (T5-3B)~\cite{gu2022don} & 75.8 & 83.4 & - & - & - & - & - & - \\ \midrule
TIARA + \textbf{GAIN} (T5-base) & 77.1 & 83.5 & 89.0 & 91.9 & 68.6 & 75.5 & 75.4 & 83.2 \\
TIARA + \textbf{GAIN} (T5-3B) & \textbf{77.1} & \textbf{83.8} & \textbf{89.0} & \textbf{92.1} & \textbf{68.8} & 76.1 & \textbf{75.4} & \textbf{83.4}  \\
GPT-3.5-turbo (5-shot) & 69.7 & 74.8 & 83.0 & 85.5 & 58.7 & 64.6 & 68.6 & 74.4 \\ 
\bottomrule
\end{tabular}
\caption{EM and F1 scores (\%) on the GrailQA valid set. ELF denotes exemplary logical form \cite{shu22tiara}. }
\label{tab:grailqa validation results}
\end{table*}

\subsection{Analysis of Integrated Aspect}

\paragraph{Hard Transfer Across Datasets}

We evaluate the performance of pre-trained models on the human-curated WebQSP dataset without fine-tuning, as shown in Table~\ref{tab:webqsp dataset transfer}.
BERT+Ranking \cite{gu21beyond} and TIARA+GAIN \cite{shu22tiara} are trained on the large-scale GrailQA dataset.
We compare these results to the state-of-the-art Pangu \cite{gu2022don}, which is fine-tuned on WebQSP and achieves an F1 score of 79.6\% 
Although we recognize that GAIN and large models offer few advantages, the performance of these pre-trained models without fine-tuning is considerably lower than Pangu's.

\paragraph{Causes from Data Collection}
We attribute this to the significant differences between training and test data, as shown in Table~\ref{tab:dataset statistics}. 
The question length, the difficulty of entity/relation linking\footnote{Measured by literal similarity: \url{https://anhaidgroup.github.io/py_stringmatching/v0.3.x/PartialRatio}.}, and the proportion of unseen schema vary dramatically across KBQA datasets. 
These discrepancies arise from the dataset construction process: WebQSP is an annotation of search logs, whereas the remaining datasets are derived from graph search and crowdsourcing.
To further enhance robustness in cross-dataset transfer, we believe that better data collection methods are required to obtain diverse and balanced training data. 
Additionally, the representation of the logical form increases the transfer difficulty, as the S-expression used in the GrailQA dataset cannot express all queries in WebQSP. 

\subsection{Analysis of Modal Aspect}

\paragraph{Context Alone Is Insufficient}
We evaluate the performance of GPT-3.5 using retrieved KB contexts (\S\ref{subsec:retrieval augmentation for LLMs}) and in-context learning on the GrailQA dataset. 
The prompts for the model include the task description and the few-shot KB contexts.
As illustrated in Table~\ref{tab:grailqa validation results}, when provided with contexts from the TIARA+GAIN retrievers, GPT-3.5 outperforms two compared models but falls short compared to TIARA+GAIN. 
Among the GPT-3.5 predictions, 79.62\% come directly from the substring of the corresponding prompts, achieving an average F1 score of 86.19\% for this portion. 
However, the remaining predictions are not part of their prompts and are entirely new predictions generated by GPT-3.5, with an average F1 score of merely 30.29\%. 
Although a baseline level is attained, these results suggest that GPT-3.5 cannot be accurately grounded to the KB environment when it does not copy the retrievers' contexts.
It also shows the modal severance of natural language pre-training and KB contexts for the LLM.
LLMs are not able to utilize KB contexts as proficiently as they understand natural language, and the faithfulness and controllability of grounded LLMs are not yet guaranteed under the current approach \cite{gu2022don}.
To mitigate this problem, alternative paradigms should be explored, such as tool learning \cite{schick23toolformer} and multi-step planning \cite{liu23agentbench} with Chain-of-Thought \cite{wei22cot}, which enables more refined access and control over environments and reduces modal differences during the reasoning process.

\begin{table}[htb]
\small
\resizebox{\linewidth}{!}{
\begin{tabular}{lrrrr}
\toprule
 & \textbf{GrailQA} & \textbf{GraphQ} & \textbf{WebQSP} & \textbf{SQB} \\ \toprule
Train size & 44,337 & 2,381 & 3,097 & 75,819 \\
Valid size   & 6,763  & -     & -     & 11,141 \\
Test size  & 13,231 & 2,395 & 1,638 & 21,483 \\ \midrule
Length     & 62.96 & 54.62 & 35.93 & 42.16  \\
\# of entities   & 0.903 & 1.028 & 1.112 & 1.000 \\
\# of relations & 1.358 & 1.434 & 1.464 & 1.000 \\ \midrule
\multicolumn{5}{c}{\textit{Similarity between questions and KB items}} \\
Entity     & 0.999 & 1.000 & 0.921 & 0.985 \\
Class      & 0.547 & 0.457 & - & - \\
Relation   & 0.470 & 0.389 & 0.300 & 0.779 \\ \midrule
\multicolumn{5}{c}{\textit{Unseen ratio (\%)}} \\
Schema     & 16.90 & 86.78 & 20.44 & 32.67 \\
Question     & 54.06 & 98.25 & 4.03 & 49.18 \\ 
\bottomrule
\end{tabular}
}\caption{KBQA dataset statistics. 
\textit{Length} denotes the average number of question characters. 
\textit{\# of entities/relations} denotes the average number of entities/relations in the logical form. 
\textit{Unseen Schema} is the ratio of unseen schema items in the dev/test set.
\textit{Unseen Question} is the ratio of questions containing unseen schema.
}
\label{tab:dataset statistics}
\end{table}

\section{Conclusion}
\label{sec:conclusion}

Despite the recent progress of LM-driven models, robustness challenges posed by distribution shifts for the KBQA task are rarely discussed.
Our analyses call for further research into better evaluation protocols for grounding LMs to KBs and enhancing the robustness of environmental, linguistic, and modal aspects.
Notably, the experiments reveal that LLMs sometimes simply copy the provided prompt. 
It indicates that the existing methodologies for grounding LLMs are yet to prove their efficacy and superiority.
Future research issues include collecting more balanced environment-specific corpora and improving the LLM learning paradigms.
For the corpora, our experiments show that the data augmentation techniques deserve further research.

\section*{Limitations}
\label{sec:limitations}

1) For question generation, the verbalization process of the GAIN method relies heavily on large-scale KBQA annotations. 
The training data influence the style of generated questions, and overly complex logical forms (e.g., with three or more hops) are difficult to convert into natural language questions. 
Besides, synthetic data is less diverse and natural than human annotations, though it improves generalization performance.
2) Multilingual KBQA is a problem that lacks attention in the KBQA research and is also a linguistic-aspect challenge.  However, since most KBQA datasets are based on English and do not discuss other languages at all, this paper leaves the evaluation of this problem for future work.

\section*{Ethics Statement}
\label{sec:ethics statement}

The proposed data augmentation method GAIN could be used on any KB.
The Freebase \cite{bollacker08freebase} used in this work is a KB that has been publicly released and manually reviewed.
For uncensored KBs, if harmful information is collected, it could make synthetic data contain harmful information and make LMs generate harmful answers.

\section*{Acknowledgments}

We extend our sincere gratitude to all the anonymous reviewers for their insightful suggestions.
Special thanks to Xiao Xu, Xiang Huang, and Sitao Cheng for their valuable discussions on this paper. 
We appreciate Yu Gu's efforts in evaluating our submissions on the GrailQA benchmark test set.

\bibliography{anthology,custom}

\begin{thebibliography}{52}
\expandafter\ifx\csname natexlab\endcsname\relax\def\natexlab#1{#1}\fi

\bibitem[{Berant et~al.(2013)Berant, Chou, Frostig, and
  Liang}]{berant13semantic}
Jonathan Berant, Andrew Chou, Roy Frostig, and Percy Liang. 2013.
\newblock \href {https://aclanthology.org/D13-1160} {Semantic parsing on
  {F}reebase from question-answer pairs}.
\newblock In \emph{Proceedings of the 2013 Conference on Empirical Methods in
  Natural Language Processing}, pages 1533--1544, Seattle, Washington, USA.
  Association for Computational Linguistics.

\bibitem[{Bi et~al.(2020)Bi, Cheng, Li, Wang, and Qi}]{bi20knowledge}
Sheng Bi, Xiya Cheng, Yuan{-}Fang Li, Yongzhen Wang, and Guilin Qi. 2020.
\newblock \href {https://doi.org/10.18653/V1/2020.COLING-MAIN.250}
  {Knowledge-enriched, type-constrained and grammar-guided question generation
  over knowledge bases}.
\newblock In \emph{Proceedings of the 28th International Conference on
  Computational Linguistics, {COLING} 2020, Barcelona, Spain (Online), December
  8-13, 2020}, pages 2776--2786. International Committee on Computational
  Linguistics.

\bibitem[{Bollacker et~al.(2008)Bollacker, Evans, Paritosh, Sturge, and
  Taylor}]{bollacker08freebase}
Kurt~D. Bollacker, Colin Evans, Praveen~K. Paritosh, Tim Sturge, and Jamie
  Taylor. 2008.
\newblock \href {https://doi.org/10.1145/1376616.1376746} {Freebase: a
  collaboratively created graph database for structuring human knowledge}.
\newblock In \emph{Proceedings of the {ACM} {SIGMOD} International Conference
  on Management of Data, {SIGMOD} 2008, Vancouver, BC, Canada, June 10-12,
  2008}, pages 1247--1250. {ACM}.

\bibitem[{Bordes et~al.(2015)Bordes, Usunier, Chopra, and
  Weston}]{bordes15large}
Antoine Bordes, Nicolas Usunier, Sumit Chopra, and Jason Weston. 2015.
\newblock \href {http://arxiv.org/abs/1506.02075} {Large-scale simple question
  answering with memory networks}.
\newblock \emph{arXiv preprint arXiv:1506.02075}.

\bibitem[{Cao et~al.(2022{\natexlab{a}})Cao, Shi, Pan, Nie, Xiang, Hou, Li, He,
  and Zhang}]{cao22kqa}
Shulin Cao, Jiaxin Shi, Liangming Pan, Lunyiu Nie, Yutong Xiang, Lei Hou,
  Juanzi Li, Bin He, and Hanwang Zhang. 2022{\natexlab{a}}.
\newblock \href {https://aclanthology.org/2022.acl-long.422} {{KQA} {Pro}: {A}
  dataset with explicit compositional programs for complex question answering
  over knowledge base}.
\newblock In \emph{Proceedings of the 60th Annual Meeting of the Association
  for Computational Linguistics (Volume 1: Long Papers), {ACL} 2022, Dublin,
  Ireland, May 22-27, 2022}, pages 6101--6119. Association for Computational
  Linguistics.

\bibitem[{Cao et~al.(2022{\natexlab{b}})Cao, Shi, Yao, Lv, Yu, Hou, Li, Liu,
  and Xiao}]{cao22program}
Shulin Cao, Jiaxin Shi, Zijun Yao, Xin Lv, Jifan Yu, Lei Hou, Juanzi Li,
  Zhiyuan Liu, and Jinghui Xiao. 2022{\natexlab{b}}.
\newblock \href {https://aclanthology.org/2022.acl-long.559} {Program transfer
  for answering complex questions over knowledge bases}.
\newblock In \emph{Proceedings of the 60th Annual Meeting of the Association
  for Computational Linguistics (Volume 1: Long Papers), {ACL} 2022, Dublin,
  Ireland, May 22-27, 2022}, pages 8128--8140. Association for Computational
  Linguistics.

\bibitem[{Chen et~al.(2021{\natexlab{a}})Chen, Tworek, Jun, Yuan, Ponde,
  Kaplan, Edwards, Burda, Joseph, Brockman, Ray, Puri, Krueger, Petrov, Khlaaf,
  Sastry, Mishkin, Chan, Gray, Ryder, Pavlov, Power, Kaiser, Bavarian, Winter,
  Tillet, Such, Cummings, Plappert, Chantzis, Barnes, Herbert-Voss, Guss,
  Nichol, Babuschkin, Balaji, Jain, Carr, Leike, Achiam, Misra, Morikawa,
  Radford, Knight, Brundage, Murati, Mayer, Welinder, McGrew, Amodei,
  McCandlish, Sutskever, and Zaremba}]{chen21evaluating}
Mark Chen, Jerry Tworek, Heewoo Jun, Qiming Yuan, Henrique Ponde, Jared Kaplan,
  Harrison Edwards, Yura Burda, Nicholas Joseph, Greg Brockman, Alex Ray, Raul
  Puri, Gretchen Krueger, Michael Petrov, Heidy Khlaaf, Girish Sastry, Pamela
  Mishkin, Brooke Chan, Scott Gray, Nick Ryder, Mikhail Pavlov, Alethea Power,
  Lukasz Kaiser, Mohammad Bavarian, Clemens Winter, Philippe Tillet,
  Felipe~Petroski Such, David~W. Cummings, Matthias Plappert, Fotios Chantzis,
  Elizabeth Barnes, Ariel Herbert-Voss, William~H. Guss, Alex Nichol, Igor
  Babuschkin, S.~Arun Balaji, Shantanu Jain, Andrew Carr, Jan Leike, Joshua
  Achiam, Vedant Misra, Evan Morikawa, Alec Radford, Matthew~M. Knight, Miles
  Brundage, Mira Murati, Katie Mayer, Peter Welinder, Bob McGrew, Dario Amodei,
  Sam McCandlish, Ilya Sutskever, and Wojciech Zaremba. 2021{\natexlab{a}}.
\newblock Evaluating large language models trained on code.
\newblock \emph{arXiv preprint arXiv:2107.03374}.

\bibitem[{Chen et~al.(2021{\natexlab{b}})Chen, Liu, Yu, Lin, Lou, and
  Jiang}]{chen21retrack}
Shuang Chen, Qian Liu, Zhiwei Yu, Chin{-}Yew Lin, Jian{-}Guang Lou, and Feng
  Jiang. 2021{\natexlab{b}}.
\newblock \href {https://aclanthology.org/2021.acl-demo.39} {{ReTraCk}: {A}
  flexible and efficient framework for knowledge base question answering}.
\newblock In \emph{Proceedings of the Joint Conference of the 59th Annual
  Meeting of the Association for Computational Linguistics and the 11th
  International Joint Conference on Natural Language Processing, {ACL} 2021 -
  System Demonstrations, Online, August 1-6, 2021}, pages 325--336. Association
  for Computational Linguistics.

\bibitem[{Das et~al.(2021)Das, Zaheer, Thai, Godbole, Perez, Lee, Tan,
  Polymenakos, and McCallum}]{das21case}
Rajarshi Das, Manzil Zaheer, Dung Thai, Ameya Godbole, Ethan Perez, Jay~Yoon
  Lee, Lizhen Tan, Lazaros Polymenakos, and Andrew McCallum. 2021.
\newblock \href {https://doi.org/10.18653/v1/2021.emnlp-main.755} {Case-based
  reasoning for natural language queries over knowledge bases}.
\newblock In \emph{Proceedings of the EMNLP 2021}, pages 9594--9611.
  Association for Computational Linguistics.

\bibitem[{Devlin et~al.(2019)Devlin, Chang, Lee, and Toutanova}]{devlin19bert}
Jacob Devlin, Ming{-}Wei Chang, Kenton Lee, and Kristina Toutanova. 2019.
\newblock \href {https://doi.org/10.18653/v1/n19-1423} {{BERT:} pre-training of
  deep bidirectional transformers for language understanding}.
\newblock In \emph{Proceedings of the 2019 Conference of the North American
  Chapter of the Association for Computational Linguistics: Human Language
  Technologies, {NAACL-HLT} 2019, Minneapolis, MN, USA, June 2-7, 2019, Volume
  1 (Long and Short Papers)}, pages 4171--4186. Association for Computational
  Linguistics.

\bibitem[{Dong et~al.(2017)Dong, Mallinson, Reddy, and Lapata}]{dong17learning}
Li~Dong, Jonathan Mallinson, Siva Reddy, and Mirella Lapata. 2017.
\newblock \href {https://doi.org/10.18653/v1/d17-1091} {Learning to paraphrase
  for question answering}.
\newblock In \emph{Proceedings of the 2017 Conference on Empirical Methods in
  Natural Language Processing, {EMNLP} 2017, Copenhagen, Denmark, September
  9-11, 2017}, pages 875--886. Association for Computational Linguistics.

\bibitem[{Dubey et~al.(2019)Dubey, Banerjee, Abdelkawi, and
  Lehmann}]{du19lcquad2}
Mohnish Dubey, Debayan Banerjee, Abdelrahman Abdelkawi, and Jens Lehmann. 2019.
\newblock \href {https://doi.org/10.1007/978-3-030-30796-7\_5} {{LC-QuAD 2.0}:
  {A} large dataset for complex question answering over wikidata and dbpedia}.
\newblock In \emph{The Semantic Web - {ISWC} 2019 - 18th International Semantic
  Web Conference, Auckland, New Zealand, October 26-30, 2019, Proceedings, Part
  {II}}, volume 11779 of \emph{Lecture Notes in Computer Science}, pages
  69--78. Springer.

\bibitem[{Gu et~al.(2022{\natexlab{a}})Gu, Deng, and Su}]{gu2022don}
Yu~Gu, Xiang Deng, and Yu~Su. 2022{\natexlab{a}}.
\newblock Don't generate, discriminate: A proposal for grounding language
  models to real-world environments.
\newblock \emph{arXiv preprint arXiv:2212.09736}.

\bibitem[{Gu et~al.(2021)Gu, Kase, Vanni, Sadler, Liang, Yan, and
  Su}]{gu21beyond}
Yu~Gu, Sue Kase, Michelle Vanni, Brian Sadler, Percy Liang, Xifeng Yan, and
  Yu~Su. 2021.
\newblock Beyond {I.I.D.}: three levels of generalization for question
  answering on knowledge bases.
\newblock In \emph{Proceedings of the Web Conference 2021}, pages 3477--3488.

\bibitem[{Gu et~al.(2022{\natexlab{b}})Gu, Pahuja, Cheng, and
  Su}]{gu22knowledge}
Yu~Gu, Vardaan Pahuja, Gong Cheng, and Yu~Su. 2022{\natexlab{b}}.
\newblock Knowledge base question answering: A semantic parsing perspective.
\newblock \emph{arXiv preprint arXiv:2209.04994}.

\bibitem[{Gu and Su(2022)}]{gu22arcaneqa}
Yu~Gu and Yu~Su. 2022.
\newblock \href {https://aclanthology.org/2022.coling-1.148} {{A}rcane{QA}:
  Dynamic program induction and contextualized encoding for knowledge base
  question answering}.
\newblock In \emph{Proceedings of the 29th International Conference on
  Computational Linguistics}, pages 1718--1731, Gyeongju, Republic of Korea.
  International Committee on Computational Linguistics.

\bibitem[{Guo et~al.(2022)Guo, Zhang, Wang, Zhang, Li, and Chen}]{guo22dsm}
Shasha Guo, Jing Zhang, Yanling Wang, Qianyi Zhang, Cuiping Li, and Hong Chen.
  2022.
\newblock {DSM}: Question generation over knowledge base via modeling diverse
  subgraphs with meta-learner.

\bibitem[{Hendrycks et~al.(2020)Hendrycks, Basart, Mu, Kadavath, Wang, Dorundo,
  Desai, Zhu, Parajuli, Guo, Song, Steinhardt, and Gilmer}]{hendrycks20themany}
Dan Hendrycks, Steven Basart, Norman Mu, Saurav Kadavath, Frank Wang, Evan
  Dorundo, Rahul Desai, Tyler~Lixuan Zhu, Samyak Parajuli, Mike Guo,
  Dawn~Xiaodong Song, Jacob Steinhardt, and Justin Gilmer. 2020.
\newblock \href {https://api.semanticscholar.org/CorpusID:220250257} {The many
  faces of robustness: A critical analysis of out-of-distribution
  generalization}.
\newblock \emph{2021 IEEE/CVF International Conference on Computer Vision
  (ICCV)}, pages 8320--8329.

\bibitem[{Hu et~al.(2019)Hu, Zou, and Zhu}]{hu2019question}
Sen Hu, Lei Zou, and Zhanxing Zhu. 2019.
\newblock How question generation can help question answering over knowledge
  base.
\newblock In \emph{CCF International Conference on Natural Language Processing
  and Chinese Computing}, pages 80--92. Springer.

\bibitem[{Hu et~al.(2021)Hu, Shu, Huang, and Qu}]{hu21edg}
Xixin Hu, Yiheng Shu, Xiang Huang, and Yuzhong Qu. 2021.
\newblock \href {https://doi.org/10.1007/978-3-030-88361-4\_8} {{EDG}-based
  question decomposition for complex question answering over knowledge bases}.
\newblock In \emph{Proceedings of the ISWC 2021}, volume 12922 of \emph{Lecture
  Notes in Computer Science}, pages 128--145. Springer.

\bibitem[{Hu et~al.(2022)Hu, Wu, Shu, and Qu}]{hu2022logical}
Xixin Hu, Xuan Wu, Yiheng Shu, and Yuzhong Qu. 2022.
\newblock \href {https://aclanthology.org/2022.coling-1.145} {Logical form
  generation via multi-task learning for complex question answering over
  knowledge bases}.
\newblock In \emph{Proceedings of the 29th International Conference on
  Computational Linguistics}, pages 1687--1696, Gyeongju, Republic of Korea.
  International Committee on Computational Linguistics.

\bibitem[{Huang et~al.(2023)Huang, Cheng, Shu, Bao, and Qu}]{huang23question}
Xiang Huang, Sitao Cheng, Yiheng Shu, Yuheng Bao, and Yuzhong Qu. 2023.
\newblock \href {https://doi.org/10.1609/AAAI.V37I11.26519} {Question
  decomposition tree for answering complex questions over knowledge bases}.
\newblock In \emph{Thirty-Seventh {AAAI} Conference on Artificial Intelligence,
  {AAAI} 2023, Thirty-Fifth Conference on Innovative Applications of Artificial
  Intelligence, {IAAI} 2023, Thirteenth Symposium on Educational Advances in
  Artificial Intelligence, {EAAI} 2023, Washington, DC, USA, February 7-14,
  2023}, pages 12924--12932. {AAAI} Press.

\bibitem[{Hupkes et~al.(2022)Hupkes, Giulianelli, Dankers, Artetxe, Elazar,
  Pimentel, Christodoulopoulos, Lasri, Saphra, Sinclair, Ulmer, Schottmann,
  Batsuren, Sun, Sinha, Khalatbari, Ryskina, Frieske, Cotterell, and
  Jin}]{hupkes22state}
Dieuwke Hupkes, Mario Giulianelli, Verna Dankers, Mikel Artetxe, Yanai Elazar,
  Tiago Pimentel, Christos Christodoulopoulos, Karim Lasri, Naomi Saphra,
  Arabella Sinclair, Dennis Ulmer, Florian Schottmann, Khuyagbaatar Batsuren,
  Kaiser Sun, Koustuv Sinha, Leila Khalatbari, Maria Ryskina, Rita Frieske,
  Ryan Cotterell, and Zhijing Jin. 2022.
\newblock \href {https://doi.org/10.48550/ARXIV.2210.03050} {State-of-the-art
  generalisation research in {NLP:} a taxonomy and review}.
\newblock \emph{CoRR}, abs/2210.03050.

\bibitem[{Keysers et~al.(2020)Keysers, Sch{\"{a}}rli, Scales, Buisman, Furrer,
  Kashubin, Momchev, Sinopalnikov, Stafiniak, Tihon, Tsarkov, Wang, van Zee,
  and Bousquet}]{keysers20measuring}
Daniel Keysers, Nathanael Sch{\"{a}}rli, Nathan Scales, Hylke Buisman, Daniel
  Furrer, Sergii Kashubin, Nikola Momchev, Danila Sinopalnikov, Lukasz
  Stafiniak, Tibor Tihon, Dmitry Tsarkov, Xiao Wang, Marc van Zee, and Olivier
  Bousquet. 2020.
\newblock \href {https://openreview.net/forum?id=SygcCnNKwr} {Measuring
  compositional generalization: {A} comprehensive method on realistic data}.
\newblock In \emph{{ICLR} 2020}. OpenReview.net.

\bibitem[{Lan et~al.(2022)Lan, He, Jiang, Jiang, Xin~Zhao, and
  Wen}]{lan21complex}
Yunshi Lan, Gaole He, Jinhao Jiang, Jing Jiang, Wayne Xin~Zhao, and Ji-Rong
  Wen. 2022.
\newblock \href {https://doi.org/10.1109/TKDE.2022.3223858} {Complex knowledge
  base question answering: A survey}.
\newblock \emph{IEEE Transactions on Knowledge and Data Engineering}, pages
  1--20.

\bibitem[{Li et~al.(2023)Li, Ma, Zhuang, Gu, Su, and Chen}]{li23fewshot}
Tianle Li, Xueguang Ma, Alex Zhuang, Yu~Gu, Yu~Su, and Wenhu Chen. 2023.
\newblock Few-shot in-context learning for knowledge base question answering.
\newblock \emph{arXiv preprint arXiv:2305.01750}.

\bibitem[{Lin and Och(2004)}]{lin2004automatic}
Chin-Yew Lin and Franz~Josef Och. 2004.
\newblock Automatic evaluation of machine translation quality using longest
  common subsequence and skip-bigram statistics.
\newblock In \emph{Proceedings of the 42nd Annual Meeting of the Association
  for Computational Linguistics (ACL-04)}, pages 605--612.

\bibitem[{Liu et~al.(2023)Liu, Yu, Zhang, Xu, Lei, Lai, Gu, Gu, Ding, Men,
  Yang, Zhang, Deng, Zeng, Du, Zhang, Shen, Zhang, Su, Sun, Huang, Dong, and
  Tang}]{liu23agentbench}
Xiao Liu, Hao Yu, Hanchen Zhang, Yifan Xu, Xuanyu Lei, Hanyu Lai, Yu~Gu, Yuxian
  Gu, Hangliang Ding, Kai Men, Kejuan Yang, Shudan Zhang, Xiang Deng, Aohan
  Zeng, Zhengxiao Du, Chenhui Zhang, Shengqi Shen, Tianjun Zhang, Yu~Su, Huan
  Sun, Minlie Huang, Yuxiao Dong, and Jie Tang. 2023.
\newblock \href {https://api.semanticscholar.org/CorpusID:260682249}
  {Agentbench: Evaluating llms as agents}.
\newblock \emph{arXiv preprint 2308.03688}.

\bibitem[{OpenAI(2023)}]{gpt4}
OpenAI. 2023.
\newblock {GPT-4} technical report.
\newblock \emph{arXiv preprint arXiv:2303.08774}.

\bibitem[{Ouyang et~al.(2022)Ouyang, Wu, Jiang, Almeida, Wainwright, Mishkin,
  Zhang, Agarwal, Slama, Ray et~al.}]{ouyang2022training}
Long Ouyang, Jeff Wu, Xu~Jiang, Diogo Almeida, Carroll~L Wainwright, Pamela
  Mishkin, Chong Zhang, Sandhini Agarwal, Katarina Slama, Alex Ray, et~al.
  2022.
\newblock Training language models to follow instructions with human feedback.
\newblock \emph{arXiv preprint arXiv:2203.02155}.

\bibitem[{Papineni et~al.(2002)Papineni, Roukos, Ward, and
  Zhu}]{papineni2002bleu}
Kishore Papineni, Salim Roukos, Todd Ward, and Wei-Jing Zhu. 2002.
\newblock Bleu: a method for automatic evaluation of machine translation.
\newblock In \emph{Proceedings of the 40th annual meeting of the Association
  for Computational Linguistics}, pages 311--318.

\bibitem[{Paszke et~al.(2019)Paszke, Gross, Massa, Lerer, Bradbury, Chanan,
  Killeen, Lin, Gimelshein, Antiga, Desmaison, Kopf, Yang, DeVito, Raison,
  Tejani, Chilamkurthy, Steiner, Fang, Bai, and Chintala}]{paszke19pytorch}
Adam Paszke, Sam Gross, Francisco Massa, Adam Lerer, James Bradbury, Gregory
  Chanan, Trevor Killeen, Zeming Lin, Natalia Gimelshein, Luca Antiga, Alban
  Desmaison, Andreas Kopf, Edward Yang, Zachary DeVito, Martin Raison, Alykhan
  Tejani, Sasank Chilamkurthy, Benoit Steiner, Lu~Fang, Junjie Bai, and Soumith
  Chintala. 2019.
\newblock \href
  {http://papers.neurips.cc/paper/9015-pytorch-an-imperative-style-high-performance-deep-learning-library.pdf}
  {Pytorch: An imperative style, high-performance deep learning library}.
\newblock In H.~Wallach, H.~Larochelle, A.~Beygelzimer, F.~d\textquotesingle
  Alch\'{e}-Buc, E.~Fox, and R.~Garnett, editors, \emph{Advances in Neural
  Information Processing Systems 32}, pages 8024--8035. Curran Associates, Inc.

\bibitem[{Patel et~al.(2022)Patel, Bhattamishra, Blunsom, and
  Goyal}]{patel22revisiting}
Arkil Patel, Satwik Bhattamishra, Phil Blunsom, and Navin Goyal. 2022.
\newblock \href {https://doi.org/10.18653/V1/2022.ACL-SHORT.46} {Revisiting the
  compositional generalization abilities of neural sequence models}.
\newblock In \emph{Proceedings of the 60th Annual Meeting of the Association
  for Computational Linguistics (Volume 2: Short Papers), {ACL} 2022, Dublin,
  Ireland, May 22-27, 2022}, pages 424--434. Association for Computational
  Linguistics.

\bibitem[{Perevalov et~al.(2022)Perevalov, Diefenbach, Usbeck, and
  Both}]{Perevalov22qald9}
Aleksandr Perevalov, Dennis Diefenbach, Ricardo Usbeck, and Andreas Both. 2022.
\newblock \href {https://doi.org/10.1109/ICSC52841.2022.00045} {{QALD-9-plus}:
  {A} multilingual dataset for question answering over dbpedia and wikidata
  translated by native speakers}.
\newblock In \emph{16th {IEEE} International Conference on Semantic Computing,
  {ICSC} 2022, Laguna Hills, CA, USA, January 26-28, 2022}, pages 229--234.
  {IEEE}.

\bibitem[{Raffel et~al.(2020)Raffel, Shazeer, Roberts, Lee, Narang, Matena,
  Zhou, Li, and Liu}]{raffel20exploring}
Colin Raffel, Noam Shazeer, Adam Roberts, Katherine Lee, Sharan Narang, Michael
  Matena, Yanqi Zhou, Wei Li, and Peter~J. Liu. 2020.
\newblock \href {http://jmlr.org/papers/v21/20-074.html} {Exploring the limits
  of transfer learning with a unified text-to-text transformer}.
\newblock \emph{Journal of Machine Learning Research}, 21:140:1--140:67.

\bibitem[{Reddy et~al.(2017)Reddy, T{\"a}ckstr{\"o}m, Petrov, Steedman, and
  Lapata}]{reddy17universal}
Siva Reddy, Oscar T{\"a}ckstr{\"o}m, Slav Petrov, Mark Steedman, and Mirella
  Lapata. 2017.
\newblock \href {https://doi.org/10.18653/v1/D17-1009} {Universal semantic
  parsing}.
\newblock In \emph{Proceedings of the 2017 Conference on Empirical Methods in
  Natural Language Processing}, pages 89--101, Copenhagen, Denmark. Association
  for Computational Linguistics.

\bibitem[{Robertson et~al.(2009)Robertson, Zaragoza
  et~al.}]{robertson09probabilistic}
Stephen Robertson, Hugo Zaragoza, et~al. 2009.
\newblock The probabilistic relevance framework: {BM25} and beyond.
\newblock \emph{Foundations and Trends{\textregistered} in Information
  Retrieval}, 3(4):333--389.

\bibitem[{Schick et~al.(2023)Schick, Dwivedi-Yu, Dess{\`\i}, Raileanu, Lomeli,
  Zettlemoyer, Cancedda, and Scialom}]{schick23toolformer}
Timo Schick, Jane Dwivedi-Yu, Roberto Dess{\`\i}, Roberta Raileanu, Maria
  Lomeli, Luke Zettlemoyer, Nicola Cancedda, and Thomas Scialom. 2023.
\newblock Toolformer: Language models can teach themselves to use tools.
\newblock \emph{arXiv preprint arXiv:2302.04761}.

\bibitem[{Shu et~al.(2022)Shu, Yu, Li, Karlsson, Ma, Qu, and Lin}]{shu22tiara}
Yiheng Shu, Zhiwei Yu, Yuhan Li, B{\"o}rje Karlsson, Tingting Ma, Yuzhong Qu,
  and Chin-Yew Lin. 2022.
\newblock \href {https://aclanthology.org/2022.emnlp-main.555} {{TIARA}:
  Multi-grained retrieval for robust question answering over large knowledge
  base}.
\newblock In \emph{Proceedings of the 2022 Conference on Empirical Methods in
  Natural Language Processing}, pages 8108--8121, Abu Dhabi, United Arab
  Emirates. Association for Computational Linguistics.

\bibitem[{Su(2023)}]{su2023language}
Yu~Su. 2023.
\newblock \href {https://yusu.substack.com/p/language-agents} {Language agents:
  a critical evolutionary step of artificial intelligence}.
\newblock \emph{yusu.substack.com}.

\bibitem[{Su et~al.(2016)Su, Sun, Sadler, Srivatsa, Gur, Yan, and
  Yan}]{su16graphq}
Yu~Su, Huan Sun, Brian~M. Sadler, Mudhakar Srivatsa, Izzeddin Gur, Zenghui Yan,
  and Xifeng Yan. 2016.
\newblock \href {https://doi.org/10.18653/v1/d16-1054} {On generating
  characteristic-rich question sets for {QA} evaluation}.
\newblock In \emph{Proceedings of the 2016 Conference on Empirical Methods in
  Natural Language Processing, {EMNLP} 2016, Austin, Texas, USA, November 1-4,
  2016}, pages 562--572. The Association for Computational Linguistics.

\bibitem[{Sun et~al.(2020)Sun, Zhang, Cheng, and Qu}]{sun20sparqa}
Yawei Sun, Lingling Zhang, Gong Cheng, and Yuzhong Qu. 2020.
\newblock {SPARQA}: skeleton-based semantic parsing for complex questions over
  knowledge bases.
\newblock In \emph{Proceedings of the AAAI Conference on Artificial
  Intelligence}, volume~34, pages 8952--8959.

\bibitem[{Talmor and Berant(2018)}]{talmor2018web}
Alon Talmor and Jonathan Berant. 2018.
\newblock The web as a knowledge-base for answering complex questions.
\newblock In \emph{Proceedings of the 2018 Conference of the North American
  Chapter of the Association for Computational Linguistics: Human Language
  Technologies, Volume 1 (Long Papers)}, pages 641--651.

\bibitem[{Touvron et~al.(2023)Touvron, Martin, Stone, Albert, Almahairi,
  Babaei, Bashlykov, Batra, Bhargava, Bhosale, Bikel, Blecher, Ferrer, Chen,
  Cucurull, Esiobu, Fernandes, Fu, Fu, Fuller, Gao, Goswami, Goyal, Hartshorn,
  Hosseini, Hou, Inan, Kardas, Kerkez, Khabsa, Kloumann, Korenev, Koura,
  Lachaux, Lavril, Lee, Liskovich, Lu, Mao, Martinet, Mihaylov, Mishra,
  Molybog, Nie, Poulton, Reizenstein, Rungta, Saladi, Schelten, Silva, Smith,
  Subramanian, Tan, Tang, Taylor, Williams, Kuan, Xu, Yan, Zarov, Zhang, Fan,
  Kambadur, Narang, Rodriguez, Stojnic, Edunov, and Scialom}]{touvron23Llama2}
Hugo Touvron, Louis Martin, Kevin~R. Stone, Peter Albert, Amjad Almahairi,
  Yasmine Babaei, Nikolay Bashlykov, Soumya Batra, Prajjwal Bhargava, Shruti
  Bhosale, Daniel~M. Bikel, Lukas Blecher, Cristian~Cant{\'o}n Ferrer, Moya
  Chen, Guillem Cucurull, David Esiobu, Jude Fernandes, Jeremy Fu, Wenyin Fu,
  Brian Fuller, Cynthia Gao, Vedanuj Goswami, Naman Goyal, Anthony~S.
  Hartshorn, Saghar Hosseini, Rui Hou, Hakan Inan, Marcin Kardas, Viktor
  Kerkez, Madian Khabsa, Isabel~M. Kloumann, A.~V. Korenev, Punit~Singh Koura,
  Marie-Anne Lachaux, Thibaut Lavril, Jenya Lee, Diana Liskovich, Yinghai Lu,
  Yuning Mao, Xavier Martinet, Todor Mihaylov, Pushkar Mishra, Igor Molybog,
  Yixin Nie, Andrew Poulton, Jeremy Reizenstein, Rashi Rungta, Kalyan Saladi,
  Alan Schelten, Ruan Silva, Eric~Michael Smith, R.~Subramanian, Xia Tan, Binh
  Tang, Ross Taylor, Adina Williams, Jian~Xiang Kuan, Puxin Xu, Zhengxu Yan,
  Iliyan Zarov, Yuchen Zhang, Angela Fan, Melanie Kambadur, Sharan Narang,
  Aurelien Rodriguez, Robert Stojnic, Sergey Edunov, and Thomas Scialom. 2023.
\newblock \href {https://api.semanticscholar.org/CorpusID:259950998} {{Llama}
  2: Open foundation and fine-tuned chat models}.
\newblock \emph{arXiv preprint 2307.09288}.

\bibitem[{Trivedi et~al.(2017)Trivedi, Maheshwari, Dubey, and
  Lehmann}]{trivedi17lcquad}
Priyansh Trivedi, Gaurav Maheshwari, Mohnish Dubey, and Jens Lehmann. 2017.
\newblock \href {https://doi.org/10.1007/978-3-319-68204-4\_22} {{LC-QuAD}: {A}
  corpus for complex question answering over knowledge graphs}.
\newblock In \emph{The Semantic Web - {ISWC} 2017 - 16th International Semantic
  Web Conference, Vienna, Austria, October 21-25, 2017, Proceedings, Part
  {II}}, volume 10588 of \emph{Lecture Notes in Computer Science}, pages
  210--218. Springer.

\bibitem[{Vrandecic and Kr{\"{o}}tzsch(2014)}]{vrandecic14wikidata}
Denny Vrandecic and Markus Kr{\"{o}}tzsch. 2014.
\newblock \href {https://doi.org/10.1145/2629489} {Wikidata: a free
  collaborative knowledgebase}.
\newblock \emph{Commun. {ACM}}, 57(10):78--85.

\bibitem[{Wei et~al.(2022)Wei, Wang, Schuurmans, Bosma, Ichter, Xia, Chi, Le,
  and Zhou}]{wei22cot}
Jason Wei, Xuezhi Wang, Dale Schuurmans, Maarten Bosma, Brian Ichter, Fei Xia,
  Ed~H. Chi, Quoc~V. Le, and Denny Zhou. 2022.
\newblock \href
  {http://papers.nips.cc/paper\_files/paper/2022/hash/9d5609613524ecf4f15af0f7b31abca4-Abstract-Conference.html}
  {Chain-of-thought prompting elicits reasoning in large language models}.
\newblock In \emph{NeurIPS}.

\bibitem[{Wu et~al.(2019)Wu, Huang, Weng, Zheng, Zhang, Yan, and
  Chen}]{wu19learning}
Peng Wu, Shujian Huang, Rongxiang Weng, Zaixiang Zheng, Jianbing Zhang, Xiaohui
  Yan, and Jiajun Chen. 2019.
\newblock \href {https://doi.org/10.18653/v1/p19-1616} {Learning representation
  mapping for relation detection in knowledge base question answering}.
\newblock In \emph{Proceedings of the 57th Conference of the Association for
  Computational Linguistics, {ACL} 2019, Florence, Italy, July 28- August 2,
  2019, Volume 1: Long Papers}, pages 6130--6139. Association for Computational
  Linguistics.

\bibitem[{Xu et~al.(2023)Xu, Xu, Wang, Liu, Zhu, and McAuley}]{xu2023small}
Canwen Xu, Yichong Xu, Shuohang Wang, Yang Liu, Chenguang Zhu, and Julian
  McAuley. 2023.
\newblock Small models are valuable plug-ins for large language models.
\newblock \emph{arXiv preprint arXiv:2305.08848}.

\bibitem[{Ye et~al.(2022)Ye, Yavuz, Hashimoto, Zhou, and Xiong}]{ye21rng}
Xi~Ye, Semih Yavuz, Kazuma Hashimoto, Yingbo Zhou, and Caiming Xiong. 2022.
\newblock \href {https://doi.org/10.18653/v1/2022.acl-long.417} {{RNG}-{KBQA}:
  Generation augmented iterative ranking for knowledge base question
  answering}.
\newblock In \emph{Proceedings of the 60th Annual Meeting of the Association
  for Computational Linguistics (Volume 1: Long Papers)}, pages 6032--6043,
  Dublin, Ireland. Association for Computational Linguistics.

\bibitem[{Yih et~al.(2016)Yih, Richardson, Meek, Chang, and
  Suh}]{yih16thevalue}
Wen{-}tau Yih, Matthew Richardson, Christopher Meek, Ming{-}Wei Chang, and Jina
  Suh. 2016.
\newblock \href {https://doi.org/10.18653/v1/p16-2033} {The value of semantic
  parse labeling for knowledge base question answering}.
\newblock In \emph{Proceedings of the 54th Annual Meeting of the Association
  for Computational Linguistics, {ACL} 2016, August 7-12, 2016, Berlin,
  Germany, Volume 2: Short Papers}. The Association for Computer Linguistics.

\bibitem[{Yu et~al.(2022)Yu, Zhang, Ng, Zhu, Li, Wang, Hu, Wang, Wang, and
  Xiang}]{yu22decaf}
Donghan Yu, Sheng Zhang, Patrick Ng, Henghui Zhu, Alexander~Hanbo Li, Jun Wang,
  Yiqun Hu, William Wang, Zhiguo Wang, and Bing Xiang. 2022.
\newblock \href {https://doi.org/10.48550/arXiv.2210.00063} {{DecAF}: Joint
  decoding of answers and logical forms for question answering over knowledge
  bases}.
\newblock \emph{arXiv preprint arXiv:2210.00063}.

\end{thebibliography}

\appendix
\clearpage

\begin{table}[htb]
\centering
\small
\resizebox{\linewidth}{!}{
\begin{tabular}{lcccc}
\toprule
\textbf{Model of Class Retrieval} & \textbf{All} & \textbf{I.I.D.} & \textbf{Comp.} & \textbf{Zero.}\tabularnewline \midrule
ReTraCk~\cite{chen21retrack} & 94.3 & 98.1 & 97.5 & 91.3 \tabularnewline
TIARA~\cite{shu22tiara} & 95.8 & \textbf{99.6} & 97.9 & 93.4 \tabularnewline
TIARA + \textbf{GAIN}  & \textbf{96.1} & 99.6 & \textbf{98.1} & \textbf{93.8} \tabularnewline
\midrule
\textbf{Model of Relation Retrieval} & \textbf{All} & \textbf{I.I.D.} & \textbf{Comp.} & \textbf{Zero.} \tabularnewline \midrule
ReTraCk~\cite{chen21retrack} & 88.4 & 95.3 & 91.0 & 84.3 \tabularnewline
TIARA~\cite{shu22tiara} & 92.0 & 97.9 & 93.7 & 88.7 \tabularnewline
TIARA + \textbf{GAIN} & \textbf{93.0} & \textbf{99.2} & \textbf{94.1} & \textbf{89.8} \tabularnewline
\bottomrule
\end{tabular}
}
\caption{Recall (\%) of top-10 retrieved schema items on the GrailQA valid set. \textit{comp.} and \textit{zero.} denote compositional and zero-shot generalization, respectively. Note that ReTraCk uses 100 classes and 150 relations for each question, while TIARA uses 10 classes and 10 relations.}
\label{tab:grailqa schema retrieval performance}
\end{table}

\begin{table}[htb]
\centering
\small
\begin{tabular}{lcc}
\toprule
\textbf{Model on GraphQuestions} & \textbf{Class} & \textbf{Relation} \tabularnewline \midrule
TIARA$^\clubsuit$ \cite{shu22tiara}  & 81.5 & 67.1 \tabularnewline
TIARA + \textbf{GAIN} & \textbf{83.3} & \textbf{74.3} \tabularnewline
\bottomrule
\end{tabular}
\caption{Recall (\%) of top-10 retrieved schema items on the GraphQuestions test set (Freebase 2015-08-09 version). 
TIARA uses 10 classes and 10 relations for each question. $^\clubsuit$ denotes our replication results.}
\label{tab:graphq schema retrieval performance}
\end{table}

\section{Example of LLM Prompt}
\label{sec:examples of llm prompt}

We present an example of an excerpted prompt, which is shown in Figures~\ref{fig:prompt example 1} and \ref{fig:prompt example 2}. 
In particular, Figure~\ref{fig:prompt example 1} illustrates the task instruction and teaching example segment, while Figure~\ref{fig:prompt example 2} displays the input query segment.

\begin{table}[t]
\small
\centering
\resizebox{\linewidth}{!}{
\begin{tabular}{llll}
\toprule
\textbf{\#question} & \textbf{\#one-hop} & \textbf{\#two-hop} & \textbf{\#domain} \\ \midrule
127,329 & 78,668 & 48,661 & 759 \\ \midrule
\textbf{\#none} & \textbf{\#count} & \textbf{\#comparatives} & \textbf{\#superlatives} \\ \midrule
115,221 & 7,115 & 1,874 & 3,119 \\ \midrule
\textbf{\#class} & \textbf{\#relation} & \textbf{\#entity} & \\ \midrule
5,078 & 12,942 & 46,645 & \\ \bottomrule
\end{tabular}
}\caption{Statistics for the synthetic dataset of logical forms. \textit{none} denotes no function.}
\label{tab:statistics for synthetic of LF}
\end{table}

\begin{table}[t]
\small
\centering
\begin{tabular}{llll}
\toprule
\textbf{\#question} & \textbf{\#relation} & \textbf{\#subject} & \textbf{\#domain} \\ \midrule
162,557 & 7,349 & 108,804 & 673 \\ \bottomrule
\end{tabular}
\caption{Statistics for the synthetic dataset of triples. \textit{Subject} denotes subject entities.}
\label{tab:statistics for synthetic of triple}
\end{table}

\section{Details of Synthetic Data}
\label{sec:examples of synthetic data}

The graph search process of GAIN is slightly different for logical forms and triples.

\paragraph{Searching Logical Forms}
GAIN employs a graph search approach similar to GraphQuestions \cite{su16graphq} to collect logical forms derived from graph queries. 
The graph query construction process consists of four steps:
1) query template construction,
2) aligning template nodes,
3) configuring functions, and
4) verification by execution.
Query templates, obtained through random graph searching, contain nodes that represent entity/literal types (not aligned to a value), referred to as \textit{template nodes}. 
Each unaligned node in the query template is then aligned with a topic entity or literal to generate multiple aligned graph queries. 
To synthesize counting, comparative, and superlative questions and enhance the diversity of synthetic data, we add functions like \texttt{COUNT}, \texttt{ARGMIN}/\texttt{ARGMAX} \cite{gu21beyond} to graph queries. 
Since KBQA research typically assumes that questions can be answered by the KB, we execute all resulting graph queries after the above steps and remove those with a null\footnote{\texttt{Null} for querying entities and \texttt{zero} for counting.} result.

\paragraph{Searching Triples}

A single KB triple can be treated as a QA pair, where the head entity and relation together form the query, and the tail entity is the answer. 
The triple search process consists of two steps:
1) candidate relation selection, and
2) triple sampling.
First, arbitrary relations $\mathcal{R}$ are selected from the KB, without being restricted to any particular KBQA dataset. 
Then, triples are collected from head entities $\mathcal{H}$, where entities in $\mathcal{H}$ are connected to relations in $\mathcal{R}$.

\paragraph{Examples of Synthetic Data}
We present some examples of synthetic data in Table~\ref{tab:synthetic example}, where the logical form contains a variety of functions.

\paragraph{Statistics}

The statistics for KBQA datasets, including our synthetic dataset, are shown in Table~\ref{tab:kbqa datasets statistics}.
To calculate the number of domains in Table~\ref{tab:statistics for synthetic of LF}, \ref{tab:statistics for synthetic of triple}, we take the domain of each Freebase class from its first segment, except for classes starting with ``base'', where we take the first two segments, e.g., domain ``music'' for the class ``music.performance\_venue'', and domain ``base.plants'' for the class ``base.plants.plant''.

\begin{table}[ht]
\small
\centering
\resizebox{\linewidth}{!}{
\begin{tabular}{llll}
\toprule
\textbf{Metrics} & \textbf{GrailQA Dev} & \textbf{GraphQ Test} & \textbf{SQB Dev} \\ \toprule
\textbf{BLEU-4}    & 0.347 & 0.178 & 0.369 \\
\textbf{ROUGE-L} & 0.526 & 0.411 & 0.640 \\ \bottomrule
\end{tabular}
}
\caption{The performance of the question generator on KBQA datasets. 
The generator for logical form is evaluated on the GrailQA valid set and GraphQuestions test set.
The generator for triple is evaluated on the SQB valid set.}
\label{tab:qg performance}
\end{table}

\begin{table*}[htb]
\small
\resizebox{\linewidth}{!}{%
\begin{tabular}{ll}
\toprule
\textbf{Logical Form (S-expression)} & \textbf{Question} \\ \midrule
\begin{tabular}[c]{@{}l@{}}1. (AND book.journal (JOIN book.periodical.editorial\_staff\\  (AND (JOIN book.editorial\_tenure.editor m.012z2ncg) \\ (JOIN book.editorial\_tenure.title m.02h6676))))\\ (GrailQA valid set)\\ \\ 2. (AND book.journal (JOIN book.periodical.editorial\_staff \\ (AND (JOIN book.editorial\_tenure.editor m.05ws\_t6) \\ (JOIN book.editorial\_tenure.title m.02wk2cy))))\\ (GrailQA valid set)\end{tabular} & \begin{tabular}[c]{@{}l@{}}1. john oliver la gorce was the editor on the editor for what journal?\\ \\ \\ \\ \\ 2. with which journal did don slater serve as editor on the editor in chief?\end{tabular} \\ \midrule
\begin{tabular}[c]{@{}l@{}}All four S-expressions are (COUNT (AND book.reviewed\_work \\ (JOIN book.reviewed\_work.reviews\_of\_this\_work m.0240y2))) \\ (GraphQuestions training set)\end{tabular} & \begin{tabular}[c]{@{}l@{}}1. how many works did fresh air review?\\ 2. how many works were reviewed by fresh air in total?\\ 3. what is the total amount of works reviewed by fresh air?\\ 4. fresh air has reviewed how many different works?\end{tabular} \\ \bottomrule
\end{tabular}
}
\caption{Examples of paraphrases in GrailQA and GraphQuestions. 
}
\label{tab:paraphrase example}
\end{table*}

\begin{table*}[htb]
\small
\resizebox{\textwidth}{!}{
\begin{tabular}{ll}
\toprule
\textbf{Sampled Logical Form} & \textbf{Synthetic Question} \\ \toprule
\begin{tabular}[c]{@{}l@{}}(COUNT (AND people.profession (JOIN \\ people.profession.people\_with\_this\_profession m.012d40)))\\ |entity|m.012d40 jackie chan\end{tabular} 
& how many professions does jackie chan have? \\ \midrule

(AND food.beer (le food.beer.original\_gravity 1.067\textasciicircum{}\textasciicircum{}float)) 
& which beer has an original gravity less than or equal to 1.067?  \\ \midrule

\begin{tabular}[c]{@{}l@{}} (AND medicine.manufactured\_drug\_form (AND (lt \\ medicine.manufactured\_drug\_form.size 10.0\textasciicircum{}\textasciicircum{}float) \\(JOIN medicine.manufactured\_drug\_form.fda\_otc\_part m.0h9yt7z))) \\ |entity|m.0h9yt7z fda otc monograph part 348 \end{tabular} 
& \begin{tabular}[c]{@{}l@{}} which manufactured drug form has part \\ fda otc monograph part 348 and has a size smaller than 10.0? \end{tabular} \\ \midrule

\begin{tabular}[c]{@{}l@{}} (ARGMAX (AND measurement\_unit.power\_unit \\ (JOIN measurement\_unit.power\_unit.measurement\_system m.07y37)) \\ measurement\_unit.power\_unit.power\_in\_watts) \\ |entity|m.07y37 us customary units
\end{tabular}
& what is the largest power unit in the us customary units? \\ \midrule

\begin{tabular}[c]{@{}l@{}} 
(AND music.release (AND (JOIN music.release.engineers m.011mbx12) \\ (JOIN music.release.label m.0g12fn3))) \\ |entity|m.011mbx12 raynard glass |m.0g12fn3 hostyle gospel ministries \end{tabular}
& \begin{tabular}[c]{@{}l@{}}  what musical release is engineered by raynard glass and \\ labelled hostyle gospel ministries? \end{tabular} \\ \toprule

\textbf{Sampled Triple} & \textbf{Synthetic Question} \\ \toprule

\begin{tabular}[c]{@{}l@{}}D.W. Robertson, Jr. (m.09ggymq), \\ people.person.place\_of\_birth, \\ Washington, D.C. (m.0rh6k)\end{tabular} 
& where was D. W. Robertson, Jr. born \\ \midrule

\begin{tabular}[c]{@{}l@{}}Alfred Chao (m.046cmd8), \\ computer.operating\_system\_developer.operating\_systems\_developed, \\ pSOS (m.0lscq)\end{tabular} 
& what operating system did Alfred Chao develop? \\ \bottomrule
\end{tabular}
}
\caption{Examples of synthetic data. 
The logical form is S-expression \cite{gu21beyond}. 
The entity label is appended to the logical form.}
\label{tab:synthetic example}
\end{table*}

\begin{table*}[htb]
\small
\centering
\begin{tabular}{lrrrr}
\toprule
\textbf{Datasets} & \textbf{\#question} & \textbf{\#class} & \textbf{\#relation} & \textbf{\#entity} \\ \toprule
GrailQA \cite{gu21beyond} & 64,331 & 1,534 & 3,720 & 32,585 \\
GraphQuestions \cite{su16graphq} & 5,166  & 506 & 596 & 376 \\
WebQSP \cite{yih16thevalue} & 4,737 & 408 & 661 & 2,593 \\
\textbf{GAIN}-synthetic & 127,329 & 5,078 & 12,942 & 46,645 \\ \bottomrule
\end{tabular}
\caption{Statistics of KBQA datasets and the GAIN-synthetic dataset.}
\label{tab:kbqa datasets statistics}
\end{table*}

\section{Experimental Details}



\subsection{Question Generation}

Training a question generation (QG) model is the second step of GAIN.
Because logical forms in the synthetic dataset do not have corresponding human-labeled natural language questions, evaluating QG on the synthetic dataset is difficult.
Existing KBQA datasets contain questions written by crowdsourced workers for logical forms or triples, so we evaluate the question generator by generating questions on these datasets and calculate the BLEU-4 \cite{papineni2002bleu} and ROUGE-L \cite{lin2004automatic} scores (Table~\ref{tab:qg performance}).\footnote{Calculated by \href{https://github.com/huggingface/evaluate}{Hugging Face Evaluate}.}

\subsection{Retrieval Performance}

The performance of schema retrieval on the GrailQA valid set and GraphQuestions test set is shown in Table~\ref{tab:grailqa schema retrieval performance} and \ref{tab:graphq schema retrieval performance}, respectively.
GAIN improves the performance of schema retrieval on both GrailQA and GraphQuestions. 
In particular, GAIN improves the relation recall@10 by 7.2 points on GraphQuestions.

\begin{table*}[htb]
\centering
\small
\begin{tabular}{lcccc}
\toprule
\textbf{Function} & \textbf{None} & \textbf{Count} & \textbf{Comparative} & \textbf{Superlative} \\ \midrule
ArcaneQA~\cite{gu22arcaneqa}      & 70.8/77.8 & 62.5/68.2 & 54.5/75.7 & 70.5/\textbf{75.6} \\
RnG-KBQA~\cite{ye21rng}       & 77.5/81.8 & 73.0/77.5 & 55.1/76.0 & 13.8/22.3 \\
TIARA (T5-base)~\cite{shu22tiara} & 77.8/83.1 & 76.4/81.8 & 57.4/81.4 & 58.7/69.0 \\ 
TIARA + \textbf{GAIN} ELF only & 76.8/81.7 & 73.9/80.0 & 0.0/25.3 & 0.0/8.3  \\
TIARA + \textbf{GAIN} (T5-base) & \textbf{78.6}/84.6 & \textbf{77.7}/\textbf{83.0} & 61.7/82.3 & 69.9/73.2 \\
TIARA + \textbf{GAIN} (T5-3B) & 78.5/\textbf{84.8} & 77.3/82.5 & \textbf{63.0}/\textbf{84.5} & \textbf{70.7}/74.1 \\
GPT-3.5-turbo (5-shot) & 74.1/78.0 & 66.8/70.5 & 38.3/60.5 & 43.9/52.3 \\ 
\midrule
\textbf{\# of relations} & \textbf{1} & \textbf{2} & \textbf{3} & \textbf{4} \\ \midrule
RnG-KBQA~\cite{ye21rng} & 75.7/79.3 & 65.3/74.7 & 28.6/44.5 & \textbf{100.0}/\textbf{100.0} \\
TIARA (T5-base)~\cite{shu22tiara} & 81.2/85.6 & 64.7/75.8 & 29.3/48.5 & 50.0/83.3 \\ 
TIARA + \textbf{GAIN} ELF only & 74.0/77.8 & 56.6/67.9 & 9.9/31.0 & 0.0/33.3 \\
TIARA + \textbf{GAIN} (T5-base) & \textbf{82.4}/\textbf{87.2} & 67.0/78.0 & \textbf{38.9}/49.8 & 50.0/83.3 \\
TIARA + \textbf{GAIN} (T5-3B) & 82.0/87.2 & \textbf{68.8}/\textbf{79.0} & 37.5/\textbf{51.3} & 50.0/83.3 \\
GPT-3.5-turbo (5-shot) & 75.0/78.6 & 61.9/69.6 & 19.8/36.6 & 50.0/50.0 \\ 
\midrule
\textbf{\# of entities} & \textbf{0} & \textbf{1} & \textbf{2} & \\ \midrule 
RnG-KBQA~\cite{ye21rng} & 58.5/63.6 & 75.4/79.9 & \textbf{55.6}/\textbf{73.5} & \\
TIARA (T5-base)~\cite{shu22tiara} & 77.5/83.1  & 76.6/82.6 & 49.9/68.0 \\ 
TIARA + \textbf{GAIN} ELF only & 42.8/47.0 & 74.2/79.9 & 47.4/67.6 \\
TIARA + \textbf{GAIN} (T5-base) & \textbf{82.2}/86.5 & 77.6/83.7 & 53.4/71.4 \\
TIARA + \textbf{GAIN} (T5-3B) & 82.0/\textbf{86.6} & \textbf{77.6}/\textbf{84.0} & 55.6/73.0 \\
GPT-3.5-turbo (5-shot) & 66.8/72.8 & 71.9/76.1 & 48.8/62.1 & \\ 
\bottomrule
\end{tabular}
\caption{EM and F1 scores (\%) for different types of logical forms on the GrailQA valid set. \textit{None} denotes no function. 
\textit{\# of relations/entities} denotes the number of relations/entities in the S-expression.
ELF denotes exemplary logical form \cite{shu22tiara}.
}
\label{tab:detail analysis}
\end{table*}

\subsection{Performance on Various Logical Forms}
\label{subsec: performance on various lf}

To show how TIARA+GAIN performs on different types of logical forms, we compare it with previous KBQA models on the GrailQA valid set, as shown in Table~\ref{tab:detail analysis}.
TIARA+GAIN improves performance in nearly all these scenarios compared to TIARA.

\subsection{Error Analysis}

To analyze the QA errors of TIARA+GAIN (T5-3B), we randomly sample 50 questions where predicted logical forms are not the same as the ground truth in the GrailQA valid set.
We follow \citet{shu22tiara} in their classification of errors. 
Entity linking errors (missing or redundant entities), 
syntactic errors (generating the wrong logical form structure), semantic errors (generating the wrong schema item when correct contexts are provided), 
false negatives (flawed annotations), 
and miscellaneous (e.g., ambiguity in the question) account for 48\%, 26\%, 16\%, 4\%, and 6\%, respectively.
For entity linking errors, 62.5\% of them are from the zero-shot level.
For syntactic errors, the number of errors from zero-shot and compositional levels is almost the same.
It means that entity linking for zero-shot domains and logical form generation for complex structures remain significant challenges.

\section{Details of Scientific Artifacts}
\label{sec:details of scientific artifacts}

All datasets we use are publicly available.
GrailQA\footnote{\url{https://dki-lab.github.io/GrailQA}} \cite{gu21beyond} uses CC BY-SA 4.0 license, and GraphQuestions\footnote{\url{https://github.com/ysu1989/GraphQuestions}} \cite{su16graphq} uses CC BY 4.0 license.
WebQSP\footnote{\url{https://www.microsoft.com/en-us/download/details.aspx?id=52763}} \cite{yih16thevalue} and SimpleQuestions-Balance\footnote{\url{https://github.com/wudapeng268/KBQA-Adapter/tree/master/Data/SQB}} \cite{wu19learning} are also downloaded from their official release channels.
We have complied with their distribution rules.
These datasets involve manual construction rather than fully automated construction, which includes the review process.
They contain questions about some famous individual people, but the corresponding content is available on Freebase, which is a publicly released, human-reviewed knowledge base.

Although the training data for LLMs could be quite large, the test set annotation of GrailQA is not publicly available.
In addition, our experimental results on the test set and the validation set show the same trend, so the impact of the data contamination problem on the experiments of this paper could be ignored.

\begin{figure*}[htb]
\small
  \centering
  \begin{minipage}{\textwidth}
    \captionsetup{type=figure} 
\texttt{Given a question and Freebase contexts, write a logical form that answers the question.}\\

\texttt{Question: dark sun: wake of the ravager was designed by what video game designer?} \\
\texttt{Candidate entities:} \\
\texttt{(A) [Dark Sun: Wake of the Ravager]} \\
\texttt{Exemplary Logical Forms:} \\
\texttt{(A) (AND cvg.cvg\_designer (JOIN cvg.cvg\_designer.games\_designed [Dark Sun: Wake of the Ravager]))} \\
  \texttt{  (B) (AND cvg.cvg\_designer (JOIN (R cvg.computer\_videogame.designers) [Dark Sun: Wake of the Ravager])) }\\
  \texttt{  (C) (AND cvg.computer\_videogame (JOIN (R cvg.cvg\_designer.games\_designed) (JOIN cvg.cvg\_designer.games\_designed [Dark Sun: Wake of the Ravager])))}\\
  \texttt{  (D) (AND cvg.computer\_videogame (JOIN cvg.computer\_videogame.designers (JOIN cvg.cvg\_designer.games\_designed [Dark Sun: Wake of the Ravager])))}\\
  \texttt{  (E) (AND base.wikipedia\_infobox.video\_game (JOIN base.wikipedia\_infobox.video\_game.developer (JOIN cvg.cvg\_designer.games\_designed [Dark Sun: Wake of the Ravager])))}\\
  \texttt{  Candidate classes:}\\
  \texttt{  (A) cvg.cvg\_designer}\\
  \texttt{  (B) cvg.game\_performance}\\
  \texttt{  (C) cvg.musical\_game}\\
  \texttt{  (D) cvg.game\_character}\\
  \texttt{  (E) cvg.computer\_game\_engine\_developer}\\
  \texttt{  (F) cvg.computer\_videogame}\\
  \texttt{  (G) cvg.computer\_game\_performance\_type}\\
  \texttt{  (H) cvg.game\_version}\\
  \texttt{  (I) cvg.computer\_game\_subject}\\
  \texttt{  (J) cvg.computer\_game\_evaluation}\\
  \texttt{  Candidate relations:}\\
  \texttt{  (A) cvg.cvg\_designer.games\_designed}\\
  \texttt{  (B) cvg.computer\_videogame.designers}\\
  \texttt{  (C) cvg.computer\_videogame.prequel}\\
  \texttt{  (D) cvg.computer\_videogame.sequel}\\
  \texttt{  (E) cvg.computer\_videogame.mods}\\
  \texttt{  (F) cvg.computer\_videogame.expansions}\\
  \texttt{  (G) cvg.computer\_videogame.developer}\\
  \texttt{  (H) cvg.computer\_videogame.characters}\\
  \texttt{  (I) cvg.game\_version.game}\\
  \texttt{  (J) cvg.computer\_game\_mod.game\_modded}\\
  \texttt{  Prediction: (AND cvg.cvg\_designer (JOIN cvg.cvg\_designer.games\_designed [Dark Sun: Wake of the Ravager]))}\\
   
    \caption{Example of LLM prompt (part 1): The task instruction and $k$ teaching examples (only one is shown because of the length) with questions and their contexts.}
    \label{fig:prompt example 1}
  \end{minipage}
\end{figure*}

\begin{figure*}[htb]
\small
  \centering
  \begin{minipage}{\textwidth}
    \captionsetup{type=figure} 

\texttt{Question: worldofwarcraft is the creation of which video game designer?} \\
\texttt{Candidate entities:} \\
\texttt{(A) [worldofwarcraft]} \\
\texttt{Exemplary Logical Forms:} \\
\texttt{(A) (AND cvg.cvg\_designer (JOIN cvg.cvg\_designer.games\_designed [worldofwarcraft]))} \\
  \texttt{  (B) (AND cvg.cvg\_designer (JOIN (R cvg.computer\_videogame.designers) [worldofwarcraft])) }\\
  \texttt{  (C) (AND cvg.cvg\_designer (JOIN cvg.cvg\_designer.games\_designed (JOIN cvg.computer\_game\_expansion.expansion\_for [worldofwarcraft])))}\\
  \texttt{  (D) (AND cvg.cvg\_designer (JOIN (R cvg.computer\_videogame.designers) (JOIN cvg.computer\_game\_expansion.expansion\_for [worldofwarcraft])))}\\
  \texttt{  (E) (AND cvg.cvg\_designer (JOIN (R cvg.computer\_videogame.designers) (JOIN (R cvg.computer\_videogame.expansions) [worldofwarcraft])))}\\
  \texttt{  Candidate classes:}\\
  \texttt{  (A) games.game\_designer}\\
  \texttt{  (B) cvg.cvg\_designer}\\
  \texttt{  (C) amusement\_parks.ride\_designer}\\
  \texttt{  (D) cvg.cvg\_developer}\\
  \texttt{  (E) cvg.computer\_videogame}\\
  \texttt{  (F) cvg.computer\_game\_engine\_developer}\\
  \texttt{  (G) cvg.computer\_game\_engine}\\
  \texttt{  (H) cvg.computer\_game\_mod}\\
  \texttt{  (I) cvg.game\_performance}\\
  \texttt{  (J) cvg.musical\_game}\\
  \texttt{  Candidate relations:}\\
  \texttt{  (A) cvg.computer\_videogame.designers}\\
  \texttt{  (B) cvg.cvg\_designer.games\_designed}\\
  \texttt{  (C) games.game\_designer.games\_designed}\\
  \texttt{  (D) games.game.designer}\\
  \texttt{  (E) cvg.computer\_videogame.developer}\\
  \texttt{  (F) cvg.cvg\_developer.games\_developed}\\
  \texttt{  (G) cvg.computer\_game\_engine.developer}\\
  \texttt{  (H) cvg.computer\_videogame.expansions}\\
  \texttt{  (I) cvg.computer\_videogame.publisher}\\
  \texttt{  (J) cvg.game\_version.developer}\\
  \texttt{  Prediction: }\\
  
    \caption{Example of LLM prompt (part 2): After $k$ teaching examples, the input question and its contexts is given. The golden prediction is 
\texttt{ (AND cvg.cvg\_designer (JOIN cvg.cvg\_designer.games\_designed [worldofwarcraft]))}.}
    \label{fig:prompt example 2}
  \end{minipage}
\end{figure*}

\end{document}